\documentclass[runningheads]{llncs}

\usepackage{eccv}
\usepackage{eccvabbrv}

\usepackage{graphicx}

\usepackage[accsupp]{axessibility}  

\usepackage[dvipsnames]{xcolor}

\usepackage{comment}

\usepackage{booktabs}
\usepackage{xspace}
\usepackage{multirow}
\usepackage{amsmath}
\usepackage{amssymb}
\usepackage{pifont}
\usepackage{colortbl}
\usepackage{tablefootnote}
\usepackage{algorithm}
\usepackage{algpseudocode}
\usepackage{makecell}
\usepackage{wrapfig}
\usepackage{caption}

\newcommand{\graynote}[1]{{\color{black!50}#1}}

\newcommand{\cmark}{\ding{51}}%
\newcommand{\xmark}{\ding{55}}%

\definecolor{firstcolor}{rgb}{0.0, 0.5, 0.0}
\definecolor{secondcolor}{rgb}{0.0, 0.125, 0.5}
\definecolor{cellcolorhl}{rgb}{0.2, 1.0, 0.20}
\newcommand{\first}{{\color{firstcolor!90}\ding{182}}{ }}
\newcommand{\second}{{\color{secondcolor!90}\ding{183}}{ }}

\newcommand{\selectedcell}{\cellcolor{cellcolorhl!10}}

\definecolor{eqnewcolor}{rgb}{0.0, 0.0, 0.6}

\newcommand{\pixood}{PixOOD\xspace}%

\newcommand\DaCUPpp{DaCUP\texttt{++}\xspace}

\DeclareMathOperator*{\argmin}{arg\,min}

\algrenewcommand\alglinenumber[1]{\raisebox{.3ex}{{\color{black!25}\tiny #1\phantom{00}}}}

\usepackage{hyperref}

\usepackage{orcidlink}

\begin{document}

\title{\pixood: Pixel-Level Out-of-Distribution Detection}

\author{Tomáš Vojíř~\orcidlink{0000-0001-7324-5883} \and
Jan Šochman~\orcidlink{0000-0002-4269-8051} \and
Jiří Matas~\orcidlink{0000-0003-0863-4844}
}

\authorrunning{T.~Vojíř et al.}

\institute{   
Czech Technical University in Prague, Faculty of Electrical Engineering\\
Department of Cybernetics, Visual Recognition Group\\
\email{\{tomas.vojir, jan.sochman, matas\}@fel.cvut.cz}\\
}

\maketitle

\begin{abstract}
We propose a pixel-level out-of-distribution detection algorithm, called \pixood, which does not require training on samples
of anomalous data and is not designed for a specific application which avoids traditional training biases. In order to model
the complex intra-class variability of the in-distribution data at the pixel level, we propose an online data condensation algorithm which is more robust than standard K-means and is easily trainable through SGD.
We evaluate \pixood on a wide range of problems. It achieved state-of-the-art results on four out of seven datasets, while being competitive on the rest. 
    The source code is available at \url{https://github.com/vojirt/PixOOD}.
\end{abstract}
    
\section{Introduction}
\label{sec:intro}

\begin{wrapfigure}{r}{0.47\textwidth}
  \vspace{-2.5em}
  \centering
  \includegraphics[width=0.47\textwidth]{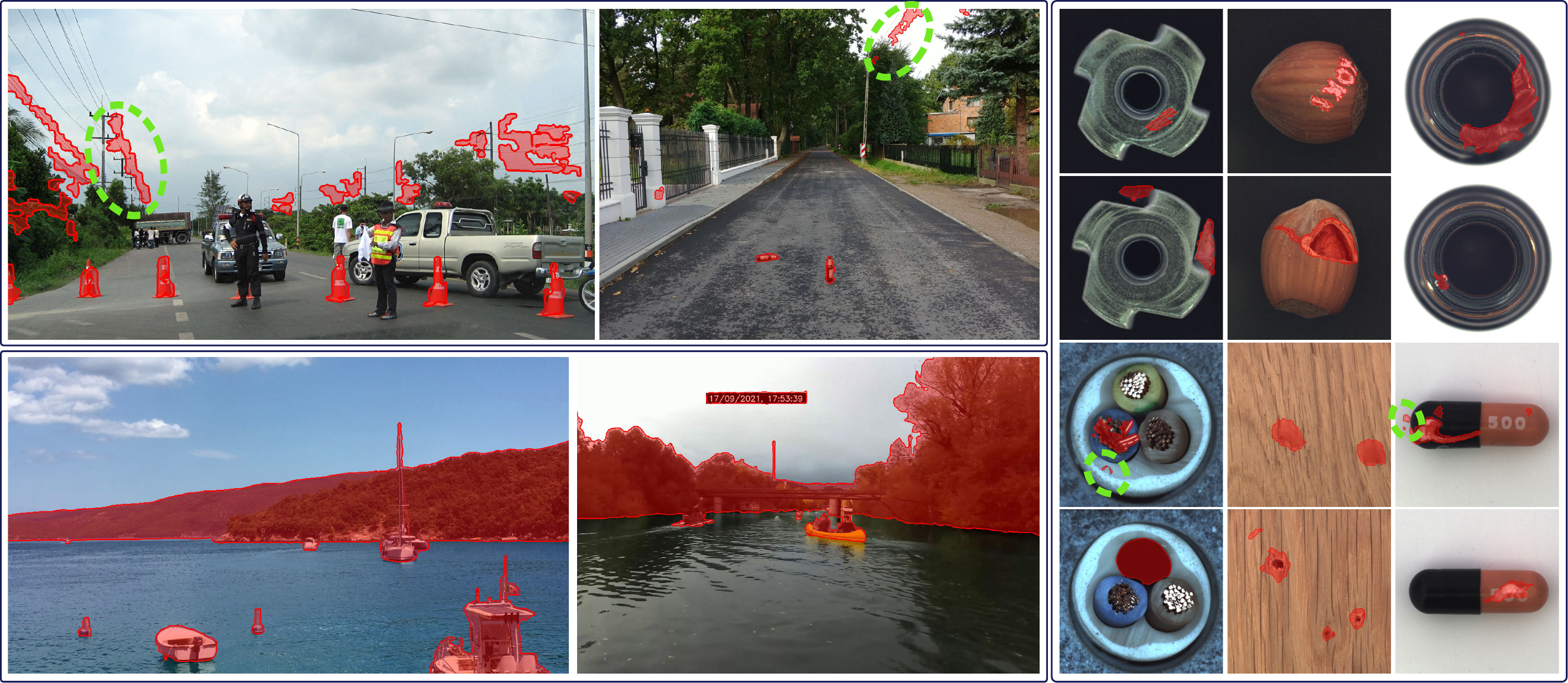}
  \caption{Examples of \pixood results for road, maritime and industrial
    anomaly detection tasks. \pixood is able to identify anomalous data not
    even considered, \ie not labelled, in standard benchmarks, \eg power cables
    (not in Cityscapes training classes) or spilled-out content or scratches in
    insulation.}
  \label{fig:intro}
  \vspace{-2em}
\end{wrapfigure}

State-of-the-art methods for many computer vision tasks rely on machine
learning and therefore on the properties of the training data; consider
segmentation, recognition, detection, optical flow, tracking, monocular depth
estimation \etc{}. In the laboratory setting, when the training and test sets
are obtained by a random split of the available corpus, the standard machine
learning assumption of identical data distribution is satisfied by
construction. However, in real-world deployment of computer vision systems,
encountering domain shifts and out-of-distribution (OOD) data is unavoidable,
as when a new product appears (recognition), new material is introduced
(segmentation) or a unique 3D structure built (monocular depth estimation).

To be robust, computer vision methods need to address the problem of recognising OOD inputs, \ie  data not
represented in the training set. Otherwise, for such data, their output is at best not
guaranteed, at worst arbitrary, potentially disastrous. Besides the potential
for reliable performance, methods that perform OOD, may benefit from
human-in-the-loop intervention on novel data, often integrating collection of OOD or rare data for annotation. 

In this paper, we present \pixood, a novel method for out-of-distribution data
detection at the level of pixels. The approach is general, it does not require
training on samples of anomalous data, nor on synthetically generated outlier
data, which requires rather specific knowledge of OOD data properties. These might be known in some case, but we aim at the scenario where minimal assumptions are made about the data not presented during training -- such data cannot be synthesized. 
The proposed method does not exploit constraints that hold in specific applications like industrial inspection
or road anomaly, or a particular problem setting like open set semantic
segmentation. \cref{fig:intro} shows examples from three diverse ODD benchmarks
\pixood is evaluated on in~\cref{sec:experiments}: the MVTec AD~\cite{Bergmann_2019_CVPR}, a batch of industrial anomaly detection
problems; the SMIYC~\cite{chan_smiyc_2021} which contains three
sub-tracks: a road anomaly detection (general anomaly segmentation in full
street scenes), obstacle detection (obstacle segmentation with the road as
region of interest), and LostAndFound; and finally the
LaRS benchmark~\cite{Zust2023LaRS}, a maritime obstacle segmentation problem
with examples of obstacles provided.

The inspiration for our method is the recently published GROOD~\cite{GROOD}
approach for image-level classification and OOD detection. We particularly
appreciate its simplicity and calibrated OOD score that is a result of solving
the Neyman-Pearson task~\cite{Neyman1928,Neyman1933} in a low dimensional projection
space. However, several shortcomings of the GROOD method limit
its applicability to diverse OOD problems. Specifically, two major issues
are (i) pixel-level generalisation (\ie decision making for each
pixel instead of the whole image) which poses several engineering challenges and
(ii) limited capability of complex intra-class variability modelling, which 
naturally arises in the pixel-level domain.

We address these weaknesses in \pixood and demonstrate its efficacy on
various downstream tasks. We show that in the case of pixel-level OOD
decision problems with large intra-class variations the GROOD single mean
representation is not sufficient. We propose a novel data condensation algorithm for
rich class appearance modelling. The condensation method is derived 
from the K-means algorithm. We show it is related to a complete data log
likelihood optimisation through the EM algorithm. The proposed method is more
robust than the standard K-means and it is easily trained on large volumes of data
through stochastic gradient descent. The condensation method is general and of interest on
its own. We also introduce several architectural improvements to
accommodate the requirements of the pixel-level decision tasks.
\noindent
The contributions of the paper are summarised as follows:
\begin{enumerate}
    \item A novel pixel-level OOD detection method called \pixood. The
        approach is general (\ie not designed for specific task/benchmark) and
        does not require any OOD training samples neither real nor synthetic (\cref{sec:method}).
    \item A novel data condensation algorithm formulated as a stochastic
        optimisation with a novel loss
        function and re-initialisation mechanism (\cref{sec:condensation}).
    \item We theoretically show the relation of the condensation loss function to a lower
        bound of the complete data log-likelihood optimisation (\cref{sec:condensation_em}).
    \item We demonstrate the applicability of the proposed method through
        applying it on three diverse benchmarks which are typically solved
        independently by specialised methods. The proposed
        method performs competitively on all (seven) datasets achieving state-of-the-art
        results on four (\cref{sec:experiments}).
\end{enumerate}

\section{Related Work}
\label{sec:related_work}

On the very basic level we divide the methods to those which use real-world or
synthetic OOD data~\cite{Tian2021PixelwiseEA, Di_Biase_2021_CVPR,
Rai_2023_ICCV, Grcic_2023_CVPR, gao2023atta, grcic22eccv, Liu_2023_ICCV,
Chan_2021_ICCV, nayal2023ICCV, Zavrtanik_2021_ICCV, cai2023} and those which do
not~\cite{liang2018enhancing, Vojir_2021_ICCV, Tian2021PixelwiseEA,
Lis_2019_ICCV, lis2023detecting, grcic2023dense, besnier2021trigger,
Vojir_2023_WACV, nayal2023ICCV, Galesso_2023_ICCV}. In our view, using a proxy
for OOD data (\eg objects from COCO in road anomaly~\cite{nayal2023ICCV, Rai_2023_ICCV} or synthetic
textures in industrial inspection~\cite{Zavrtanik_2021_ICCV}) is the same as making an assumption
about the form of allowed anomalies, which goes against the definition of the
word 'anomaly' itself. 
We aim at a method applicable to the widest possible spectrum of problems, thus
avoid introducing task specific biases through auxiliary (synthetic) OOD data.

Among the methods we compare with, one can observe several trends. In the road
anomaly detection community, the reconstructions methods~\cite{Vojir_2021_ICCV,
Vojir_2023_WACV, Lis_2019_ICCV, lis2023detecting} are trained, to
reconstruct ``normal'' pixels. The assumption is
that during inference an anomalous object would be poorly reconstructed which
could be detected. These pixel-level reconstruction-based method were slowly
overtaken by energy-based models~\cite{grcic2023dense, grcic22eccv,
Chan_2021_ICCV} that introduce regularisation to the classification objective loss
to enforce particular predictions for anomalous pixels, \eg
uniform distribution of posterior probability over in-distribution (ID) classes.
Recently, region-based methods~\cite{Rai_2023_ICCV, Grcic_2023_CVPR,
nayal2023ICCV,Zhang_Li_Qi_Yang_Ahuja_2024} gained popularity. This makes the ID/OOD reasoning more
robust to noise and makes sense in this domain since
most anomalous objects in driving scenarios have strong boundary separation from its surrounding. The
region-based methods, however, are not common in other domains such as the
industrial anomaly detection where the anomalies have different
characteristics, \eg deformations or missing parts.

In industrial anomaly benchmarks, recent successful methods are
still based on reconstruction~\cite{Zavrtanik_2021_ICCV,lu2023hierarchical,
you2022a}, likelihood modelling~\cite{Bae_2023_ICCV,
Gudovskiy_2022_WACV} or nearest-neighbour classification~\cite{Roth_2022_CVPR}.
Interestingly,  only the latest methods start to depart from a single model for each class and
start to focus on a general models~\cite{lu2023hierarchical,
you2022a}. A possible reason is that the benchmarks are rather small and
images are taken under well controlled conditions, so the generalisation is not the prime objective. 
\section{Method} \label{sec:method}
\begin{figure*}[th]
  \centering
    \includegraphics[width=0.98\linewidth]{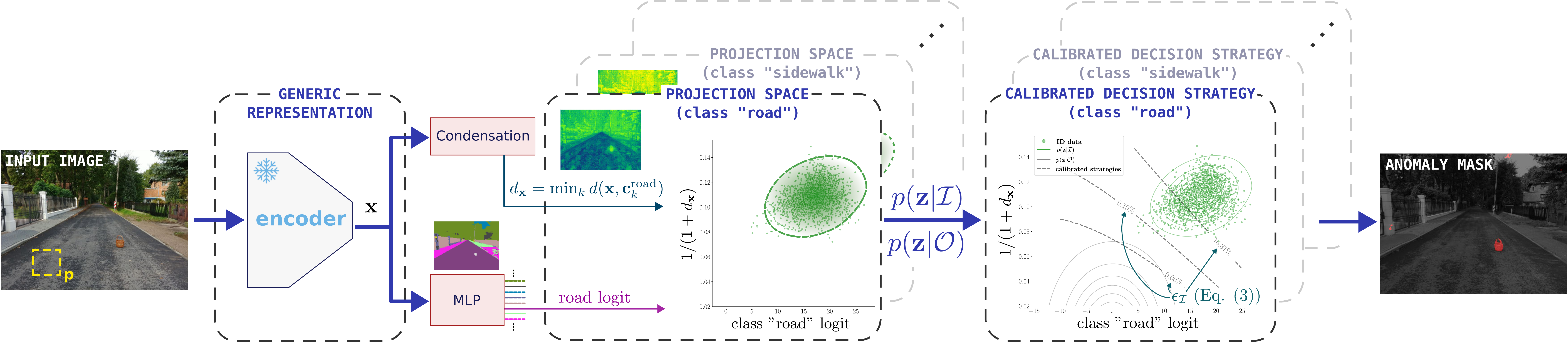} 
    \caption{\pixood inference overview. Individual steps are described
    in~\cref{sec:method}. Note that the condensation is run independently for each class during training while MLP is trained jointly for all classes.}
  \label{fig:pixood}
\end{figure*}
In this section we describe the proposed \pixood method for OOD (anomaly)
detection at the pixel level. The \pixood framework 
consists of three
components (illustrated in~\cref{fig:pixood}): (i) extraction of the
pixel/patch feature representation, (ii) building a two-dimensional projection
space, and (iii) finding the optimal and calibrated ID/OOD decision strategy.
The framework is similar to~\cite{GROOD}, which however is not applicable to pixel-level tasks due to issues discussed and addressed in~\cref{sec:im2pix}, and it is experimentally validated in~\cref{sec:ablation}.

\noindent{\bf Representation.}
Every patch $\mathbf{p}$ in the input image (the patch size, in our case, is determined
by the ViT~\cite{dosovitskiy2021an} architecture) is first transformed into
a feature vector $\mathbf{x} \in \mathbb{R}^D$ by a fixed pre-trained model.
This representation needs to be
rich enough to allow modelling ID data in a given pixel-level vision
tasks. 
A~good example of such model is the self-supervised trained DINOv2
encoder~\cite{oquab2023dinov2} that we employ in this work. Similarly,
methods such as~\cite{GROOD,gu2023anomalygpt,Wang_2023_ICCV} utilized pre-trained CLIP~\cite{radford2021learning}
image encoder for image-level OOD detection tasks.

\noindent{\bf 2D Projection Space.}
The 2D projection space serves two purposes: First, it makes it possible to estimate
the ID densities with potentially limited data (\eg~{\raise.17ex\hbox{$\scriptstyle\mathtt{\sim}$}}200 samples per class for
MVTecAD dataset), second, it allows to make reasonable assumptions about the unknown
OOD distribution, which would be difficult/impossible in the encoder's high-dimensional space.

As opposed to the image-level task~\cite{GROOD}, the
pixel/patch-level representations tend to produce more complex distributions in
the (1024-dim in our case) embedding space over an image-level semantic classes.
We observed that the simple projections used in~\cite{GROOD} is not able to model the
rich intra-class appearance variations, \eg imagine pixels (patches)
corresponding to car label may be visually 
drastically different (wheels vs.
windshield) as shown in the ablation study in~\cref{sec:experiments}. Thus we
replace the first projection, the Linear Probe, with a multi-layer
Perceptron (\cref{sec:lp}) and instead of the nearest class mean projection we
propose to use our novel data condensation algorithm~(\cref{sec:condensation}). The condensation method
is general and may find its uses also outside of the \pixood method. The
projection space is thus formed (for each class independently) as $\mathcal{Z}
= \mathcal{Z}_{\text{MLP}} \times \mathcal{Z}_{\text{N}c_k}$, where
$\mathcal{Z}_{\text{MLP}}$ is a 1-D space of MLP logit scores for the
particular class and $\mathcal{Z}_{\text{N}\mathbf{c}_k}$ is a 1-D space of
distances to the nearest class etalon found by the condensation algorithm.

\noindent{\bf Decision Strategy.} The optimal decision strategy is found independently in all these 2D projection spaces.
Let $\mathcal{I}$ be a (meta-)class representing the in-distribution
samples from a target class and $\mathcal{O}$ a (meta-)class for out-of-distribution data.
Both, $p(\mathbf{z}|\mathcal{I})$ and $p(\mathbf{z}|\mathcal{O})$ are modelled as multivariate normal
distributions in $\mathbf{z}\in\mathcal{Z}$. The parameters of $p(\mathbf{z}|\mathcal{I})$ are
estimated from the training data and $p(\mathbf{z}|\mathcal{O})$ is constructed with
zero mean and a diagonal covariance matrix with large variances
(see~\cite{GROOD} for detailed reasoning about the OOD distribution).

The ID/OOD classification problem is formulated, same as in~\cite{GROOD}, as
a Neyman-Pearson task~\cite{Neyman1928,Schlesinger2002}: Find a strategy $q^*(z): \mathcal{Z} \rightarrow
\{\mathcal{I}, \mathcal{O}\}$ such that
\begin{equation}
    \begin{aligned} 
        q^* = \argmin_q & \int_{\mathbf{z}: q(\mathbf{z}) \neq \mathcal{O}} p(\mathbf{z}|\mathcal{O})\,d\mathbf{z} \\ 
        \textrm{s.t.} \quad & \epsilon_\mathcal{I} = \int_{\mathbf{z}: q(\mathbf{z}) \neq \mathcal{I}} p(\mathbf{z}|\mathcal{I})\,d\mathbf{z} \leq \epsilon
    \end{aligned}
\end{equation}
This optimisations problem minimises the false positive rate (false acceptance
of OOD data) and bounds the ID false negative rate by~$\epsilon$.
It is known~\cite{Schlesinger2002} that the optimal strategy for a given
$\mathbf{z}\in\mathcal{Z}$ is constructed using the likelihood ratio:
\begin{equation}
        q(\mathbf{z}) = 
        \left\{
            \begin {aligned}
                 & \mathcal{I} & \text{if}\;\;\; & r(\mathbf{z}) > \mu \\
                 & \mathcal{O} & \text{if}\;\;\; & r(\mathbf{z}) \leq \mu \\
            \end{aligned}
        \right. \quad \text{where} \quad r(\mathbf{z}) = \frac{p(\mathbf{z}|\mathcal{I})}{p(\mathbf{z}|\mathcal{O})}\label{eq:opt_q}
\end{equation}
The optimal strategy $q^*$ is obtained by selecting maximal threshold $\mu$
s.t. $\epsilon_\mathcal{I} \leq \epsilon$.

\noindent{\bf ID/OOD Score.}
In practice, we would specify the acceptable false negative rate $\epsilon$ and
obtain a single optimal strategy. However, most evaluation benchmarks require
a score in the range $(0, 1)$ which is then used to compute the evaluation
metrics by varying a threshold on this score. To produce such score, we follow~\cite{GROOD} and learn
a mapping from the likelihood ratio to false negative rate. We uniformly cover 
the projection space $\mathcal{Z}$ by $M$ samples, and for each sample we find the likelihood
ratio $r(\mathbf{z}_i)$ and a corresponding error $\epsilon_\mathcal{I}^i$ for the threshold
set to $\mu = r(\mathbf{z}_i)$. This results in a set of pairs $R = \{(r(\mathbf{z}_i),
\epsilon_\mathcal{I}^i)\}_{i=1}^M$. We build an in-distribution scoring function $s_\mathcal{I}$
as a linear interpolation function over $\mathcal{Z}$ estimated from $R$ such that
\begin{equation}
    \label{eq:id_score}
    s_\mathcal{I}\left(r(\mathbf{z}_i)\right) \approx \epsilon_\mathcal{I}^i
\end{equation}
This score corresponds to the false negative error that we would make on
the in-distribution data if we selected $\mu=r(\mathbf{z})$ in the
optimal strategy $q^*$ (\cref{eq:opt_q}).

Finally, we define the OOD (``anomaly'') score as
\begin{equation}
    s_\mathcal{O}\left(r(\mathbf{z})\right) = 1 - s_\mathcal{I}\left(r(\mathbf{z})\right)
\end{equation}
The score $s_\mathcal{O}$ is used in all experiments as the output of the
\pixood method. It is in the desired $(0, 1)$ range, it is calibrated and
explainable and with a clear meaning corresponding to the false negative rate
on the ID data. See the decision boundaries in \cref{fig:pixood} for illustration of the effect of  $\epsilon_\mathcal{I}$.
\graynote{Note that the usable range of this score is at the
tail of the range, \eg $\ge 0.95$ which is equivalent to saying that the false
negative error of the ID data is $ \le 5\%$. See $\epsilon_\mathcal{I}$,
\cref{eq:id_score}, and its corresponding decisions boundaries in
\cref{fig:pixood} for illustration.}

\subsection{Incremental Soft-to-Hard Data Condensation} \label{sec:condensation}
Instead of using a single etalon for each class as in~\cite{GROOD}, the \pixood
method uses up to $K$ etalons per class. Due to the amount of data (pixels),
the etalons have to be estimated incrementally (using batches of data), ideally
in a stochastic gradient descent framework. We first motivate and formulate the
optimisation criterion for condensing a (single class) dataset into a set of
$K$ etalons. We then show that the soft-assignment condensation criterion
corresponds closely to the EM-algorithm steps when finding the optimal
parameters of a spherical Laplace distribution mixture model. This gives us
more formal view of the condensation criterion. Finally, we show how to
alleviate the issue of the local minima.

Let $T = \{\mathbf{x}_i\in \mathcal{X} \equiv \mathbb{R}^D\}_{i=1}^N$ be a (potentially large) dataset of training samples from one class. 
The dimensionality of the samples depends on the representation used (later we use $D=2$ and $D=1024$). The \textbf{dataset condensation task} is to find up to $K$ etalons (representatives) $\mathbf{c}_1, \ldots, \mathbf{c}_K$ ($\mathbf{c}_k \in \mathcal{X}$) that cover the training data, \ie all data points should be close to an etalon. The number of etalons is not fixed, but is upper-bounded by $K$ (a "budget"). Further, each etalon should represent a similar portion of the data.

Let us denote the trainable parameters, \ie the etalon coordinates, $\mathbf{c} = (\mathbf{c}_1, \ldots, \mathbf{c}_K)$
and let $d(\mathbf{x}, \mathbf{c}_k)$ be the $L_2$ distance between a data point $\mathbf{x}$ and an etalon $\mathbf{c}_k$. K-means optimises
\begin{equation}
    \mathbf{c}^* = \arg\min_\mathbf{c} \frac{1}{N}\sum_{i=1}^N \min_k d(\mathbf{x}_i, \mathbf{c}_k)^2\;.
\label{eq:kmeans}
\end{equation}
This objective is not smooth due to the hard assignment to the closest etalon. Also, K-means tends to get stuck in local minima and is known to be sensitive to outliers in data due to the square of the distance.
To make the criterion smooth, we use a smooth approximation~\cite{cho2022dkm} of the $\min$ operator:
\begin{align}
    &\mathbf{c}^* = \arg\min_\mathbf{c} \frac{1}{N}\sum_{i=1}^N \sum_{k=1}^K {\color{eqnewcolor}w(k, i; \tau)} d(\mathbf{x}_i, \mathbf{c}_k)^2 \label{eq:skmeans_tau}\\
    &w(k, i; {\color{eqnewcolor}\tau}) = \frac{\exp{\left(-d(\mathbf{x}_i, \mathbf{c}_k)^2/{\color{eqnewcolor}\tau}\right)}}{\sum_{j=1}^K \exp{\left(-d(\mathbf{x}_i, \mathbf{c}_j)^2/ {\color{eqnewcolor}\tau}\right)}}\;,\label{eq:softmin}
\end{align}
where~\cref{eq:softmin} is the softmin function with temperature scaling $\tau$. The temperature scaling allows a smooth transition from a soft assignment (easier to optimise, less susceptible to local minima) and a hard assignment (the goal of condensation).
During the optimisation, the value of $\tau$ is scheduled to 
decrease from a large (soft assignment) to a low value (hard assignment).

\begin{figure*}[t]
\centering
\captionsetup[subfigure]{labelformat=empty}
\subfloat[]{
 \setlength{\tabcolsep}{2pt}
 \tiny
  \begin{tabular}{cc}
     \includegraphics[width=0.21\textwidth]{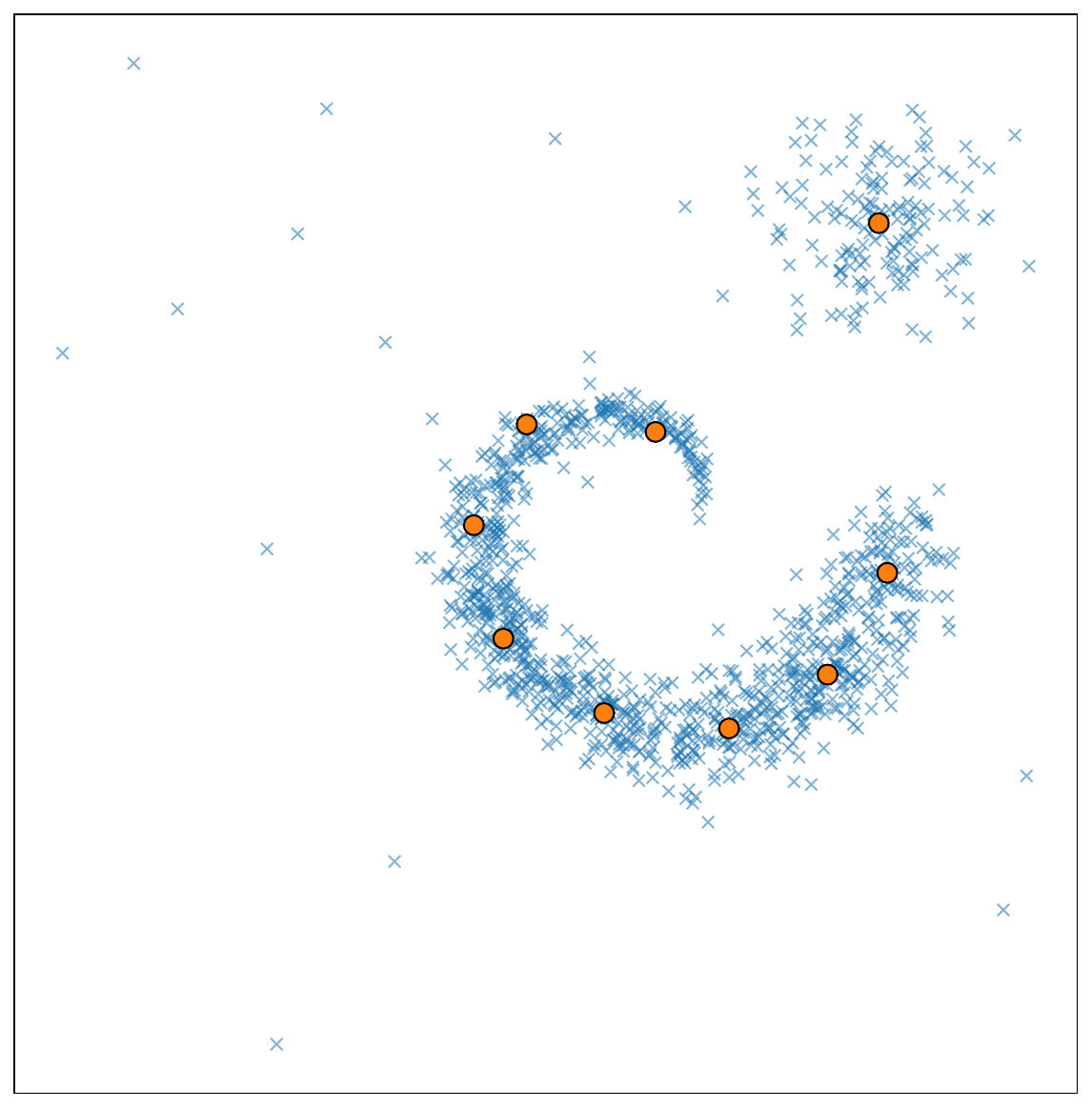} & 
      \includegraphics[width=0.21\textwidth]{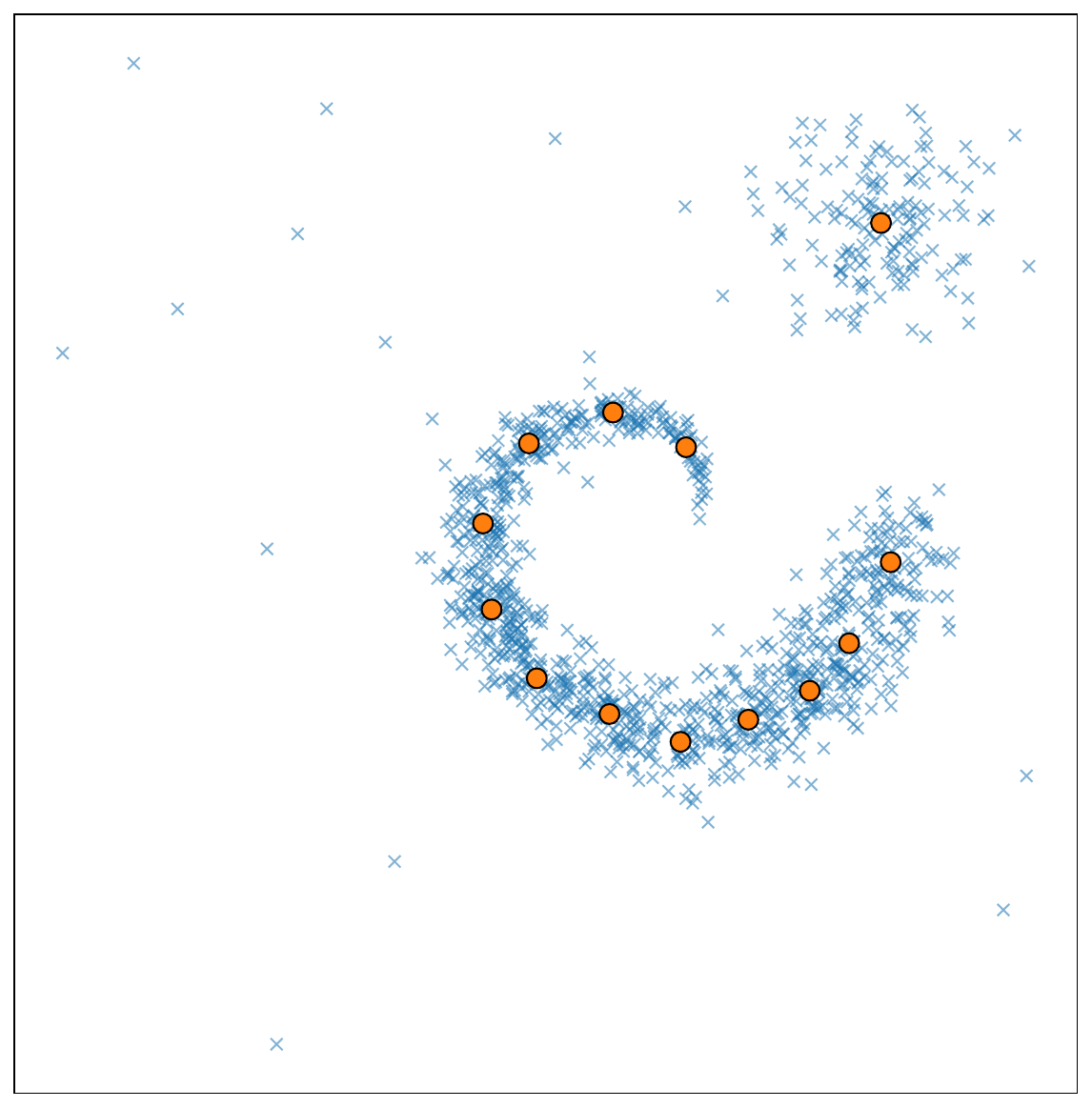}  \\
      (a) Soft K-Means, \cref{eq:skmeans_tau} & 
      (b) Soft K-Medians, \cref{eq:skmedians} \\

     \includegraphics[width=0.21\textwidth]{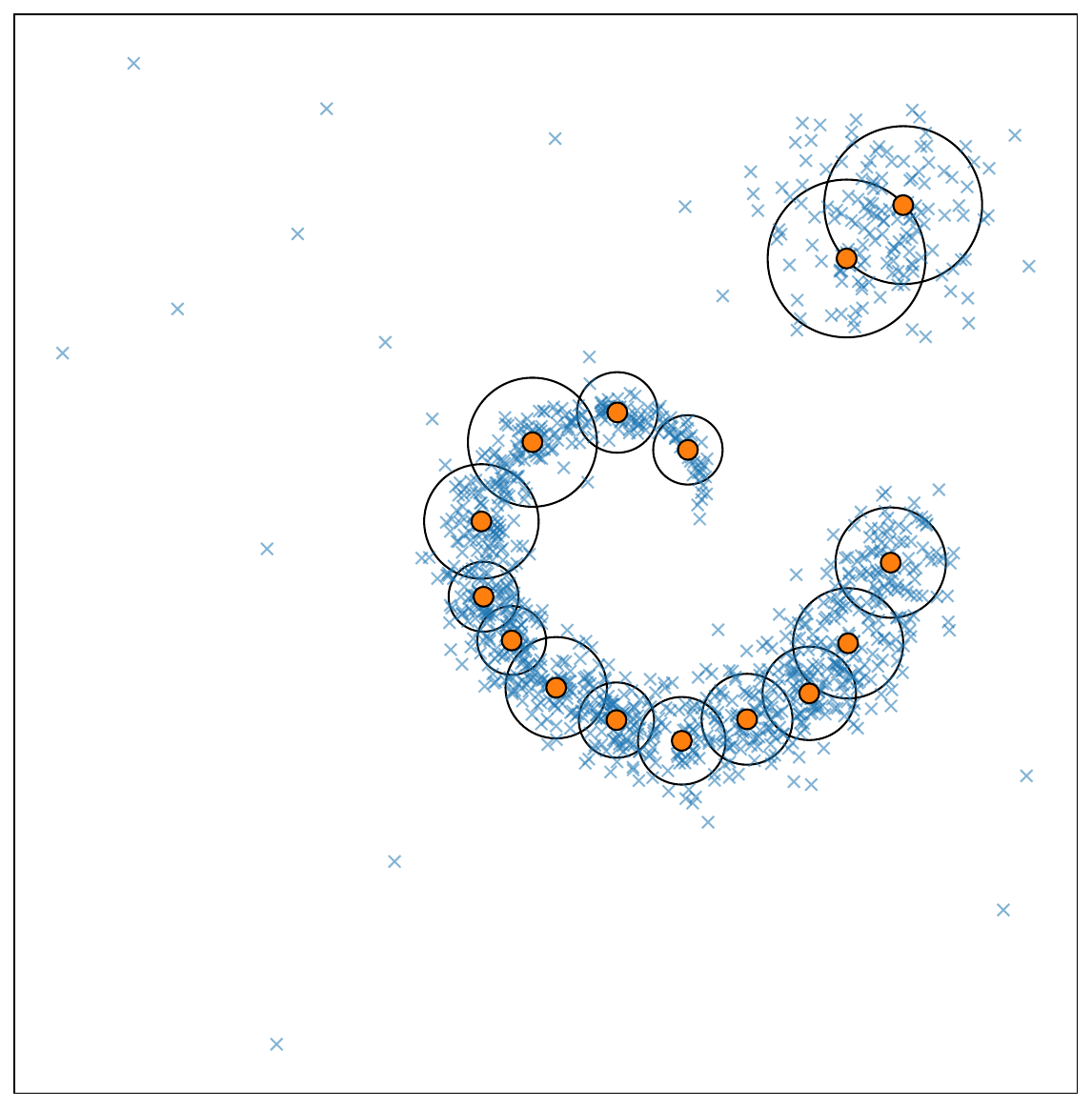} & 
      \includegraphics[width=0.21\textwidth]{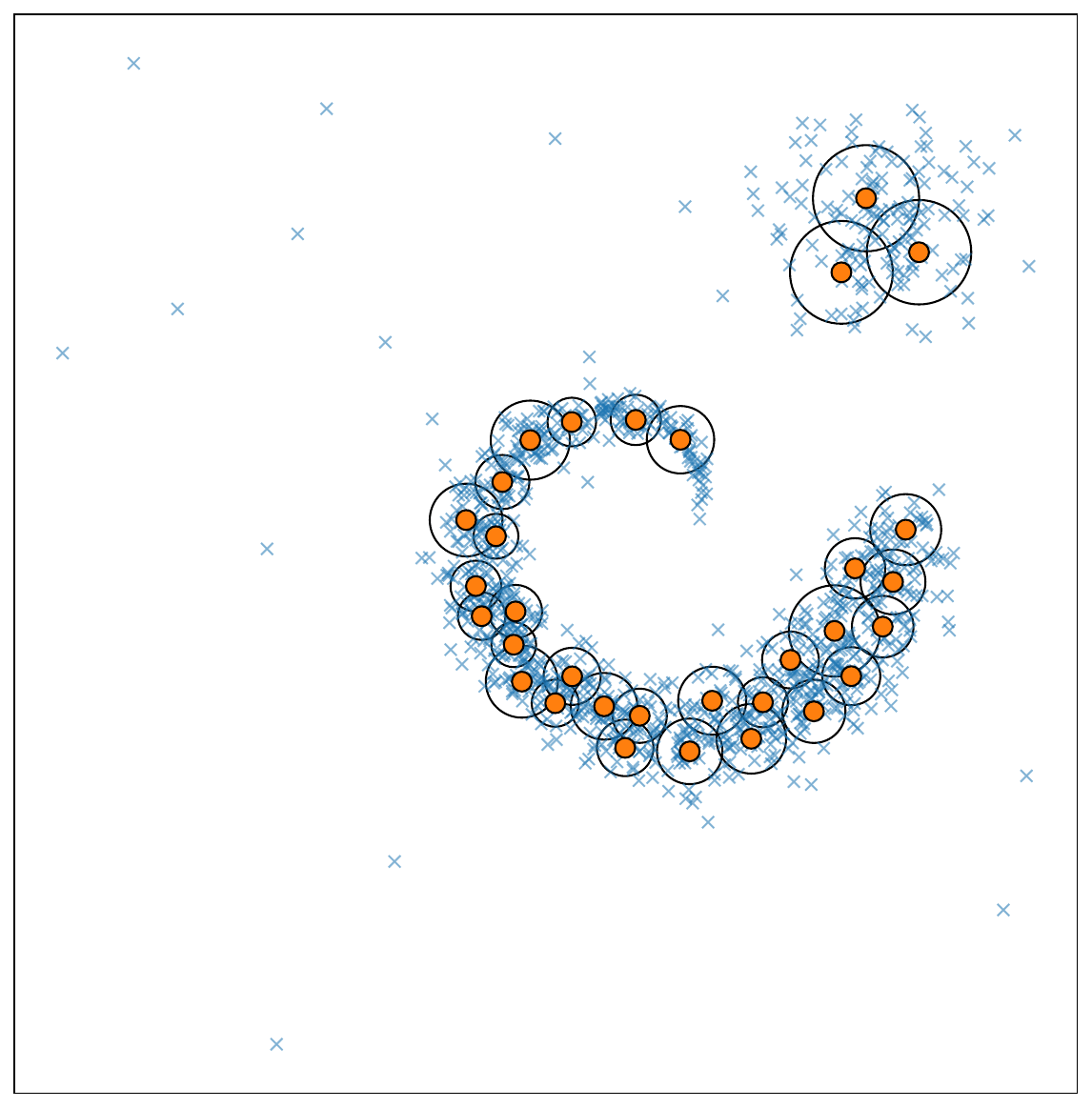} \\
      (c) Condensation, \cref{eq:condensation_loss} & 
      (d) \cref{eq:condensation_loss} + re-inits \\
  \end{tabular}
  \label{fig:cond_ablation}
}
\subfloat[]{
    \vspace{3em}
    \centering\resizebox{0.49\textwidth}{!}{
    \begin{minipage}{0.68\textwidth}
    \begin{algorithmic}[1]
    \State Given: $T = \{x_i\in \mathbb{R}^D\}_{i=1}^N$
    \State Initialise: $\mathbf{c}, \boldsymbol{\beta}, \tau, q, \lambda$
    \For{epoch $<$ num. epochs}
        \For{$X = \{\mathbf{x}_i\}_{i=1}^{n} \subset T$} \Comment{iterate over batches}
            \State Compute $d(\mathbf{x}_i, \mathbf{c}_k)\;\; \forall \mathbf{x}_i \in X, \forall \mathbf{c}_k \in \mathbf{c}$
            \State Compute $w(k, i; \tau)$ \Comment \cref{eq:wtau}\phantom{0} 
            \State Compute $L(\mathbf{c}, \boldsymbol{\beta}; T, \tau)$ \Comment \cref{eq:condensation_loss}
            \State  $\forall \mathbf{c}_k: \mathbf{c}_k \leftarrow \mathbf{c}_k - \lambda \frac{\partial L}{\partial \mathbf{c}_k}$ 
            \State  $\forall \beta_k: \beta_k \leftarrow \beta_k - \lambda \frac{\partial L}{\partial \beta_k}$
            \State Update $s_t(k)$ \Comment \cref{eq:ewa}
        \EndFor
        \If{$\text{epoch} > \text{epoch}_{\text{warmup}}$ and $s_t(k) < \theta_r$}
        \State Re-initialise $\mathbf{c}_k$
        \EndIf
        \State $\tau\leftarrow$ cos\raisebox{0.15em}{\_}scheduler(epoch)
    \EndFor
    \end{algorithmic}
    \end{minipage}
    }
    \label{alg:condensation}
}
    \caption{\textbf{Left:} Methods derived in~\cref{sec:condensation} applied
    to synthetic data (blue crosses) with outliers (isolated blue crosses).
    Only the "useful" etalons (from total of 50) with $s_t(k) >= \theta_r$ are
    displayed as orange circles. (a) K-means is sensitive to outliers and most
    etalons converge towards isolated data points (not shown) -- only 9 useful
    etalons. (b) K-medians is more robust -- 13 useful, (c) condensation adds
    the scale parameter $\beta_k$ to each cluster (black circle) enabling
    adaptive region of influence -- 15 useful, (d) re-inits with \textit{(c)}
    preserve significantly more etalons by combining re-initialization strategy
    with adaptive scale -- 32 useful. \textbf{Right:} Iterative soft-hard
    condensation algorithm summary.}
    \label{fig:cond_ablation_alg}
\end{figure*}

The smooth objective in~\cref{eq:skmeans_tau} is sensitive to outliers in data due to the square of the distance. When removing the square from~\cref{eq:kmeans}, the solution to the problem is known as the K-medians algorithm which requires an iterative optimisation for finding the medians in each step. Instead, we remove the square in the smoothed version~\cref{eq:skmeans_tau,eq:softmin}
which is still easy to optimise and allows the soft-hard assignment transition through $\tau$ parameter:
\begin{align}
    &\mathbf{c}^* = \arg\min_\mathbf{c} \frac{1}{N}\sum_{i=1}^N \sum_{k=1}^K w(k, i; \tau) {\color{eqnewcolor}d(\mathbf{x}_i, \mathbf{c}_k)} \label{eq:skmedians}\\
    &w(k, i; \tau) = \frac{\exp{(-{\color{eqnewcolor}d(\mathbf{x}_i, \mathbf{c}_k)}/\tau)}}{\sum_{j=1}^K \exp{(-{\color{eqnewcolor}d(\mathbf{x}_i, \mathbf{c}_j)}/\tau)}} \label{eq:wtau}
\end{align}
The final requirement of the condensation task is that each etalon covers
approximately the same number of data points. To that end, each etalon is
assigned a trainable scale parameter $\boldsymbol{\beta} = (\beta_1, \ldots,
\beta_K)$ to make it adaptive to the varying local data density. Instead of
just dividing the distance by the parameter ($d(\mathbf{x}_i, \mathbf{c}_k)
/ \beta_k$) in the formula, which would have a trivial minimiser for $\beta_k
= \infty$, the distance is replaced by a logarithm of the Laplace distribution
which regularises the $\beta_k$ parameter to avoid this degenerated solution.
This gives us the final loss optimised by the proposed algorithm:
\begin{equation}
    L(\mathbf{c}, \boldsymbol{\beta}; T, \tau) = -\frac{1}{N}\sum_{i=1}^N\sum_{k=1}^K w(k, i; \tau) {\color{eqnewcolor}\log \left(\frac{1}{\beta_k} e^{-\frac{d(\mathbf{x}_i, \mathbf{c}_k)}{\beta_k}}\right)} \;,
    \label{eq:condensation_loss}
\end{equation}
where $w(k, i; \tau)$ is defined in~\cref{eq:wtau}.
The full iterative soft-hard condensation algorithm is summarised
in~\cref{fig:cond_ablation_alg} (right) and the individual steps in its derivation are
illustrated on toy example in~\cref{fig:cond_ablation_alg} (left).
\subsubsection{Relation to EM Algorithm.} \label{sec:condensation_em}
Interestingly, this objective has another interpretation. It is a lower bound
on the complete data log-likelihood in EM algorithm for a mixture of spherical
Laplace distributions with equal priors.
Let us model the measurements $\mathbf{x}$ by a spherical Laplace distribution mixture model
\begin{equation}
    p(\mathbf{x}; \boldsymbol{\pi}, \boldsymbol{c}, \boldsymbol{\beta}) = \sum_{k=1}^K\pi_k p(\mathbf{x}| k; \mathbf{c}_k, \beta_k) 
    = \sum_{k=1}^K\pi_k \frac{1}{Z_k} \frac{1}{\beta_k^{D-1}} \exp{\left(-\frac{d(\mathbf{x}, \mathbf{c}_k)}{\beta_k}\right)}
\end{equation}
where $\boldsymbol{\pi}=(\pi_1, \ldots, \pi_K)$ are the prior probabilities of
each component, $\mathbf{c}=(\mathbf{c}_1, \ldots, \mathbf{c}_K)$ and
$\boldsymbol{\beta}=(\beta_1, \ldots, \beta_K)$ characterise the component
distributions as in~\cref{eq:condensation_loss} and $Z_k$'s are the normalising
factors of the spherical Laplace distribution independent of
$\boldsymbol{\beta}$ and $\mathbf{c}$. The EM maximises the complete
data log-likelihood
\begin{equation}
    \mathcal{L}(\boldsymbol{\pi}, \boldsymbol{c}, \boldsymbol{\beta}; T) = \sum_{i=1}^N \log p(\mathbf{x}; \boldsymbol{\pi}, \boldsymbol{c}, \boldsymbol{\beta})
    = \sum_{i=1}^N \log \sum_{k=1}^K\pi_k p(\mathbf{x}| k; \mathbf{c}_k, \beta_k)\;.
\end{equation}
By introducing a variational distribution $w(k, i)$ (we show how to choose it below) a lower bound to $\mathcal{L}$ is derived using Jensen's inequality as
\begin{align}
    \mathcal{L}(\boldsymbol{\pi}, \boldsymbol{c}, \boldsymbol{\beta}; \mathbf{x}) &= \sum_{i=1}^N \log \sum_{k=1}^K w(k, i) \frac{\pi_k p(\mathbf{x}| k; \mathbf{c}_k, \beta_k)}{w(k, i)}\\
    &\geq \sum_{i=1}^N \sum_{k=1}^K w(k, i) \log \frac{\pi_k p(\mathbf{x}| k; \mathbf{c}_k, \beta_k)}{w(k, i)}
\end{align}
which for the case of a mixture of spherical Laplace distributions looks like
\begin{equation}
    \sum_{i=1}^N \sum_{k=1}^K w(k, i) \log\pi_k + w(k, i) \log \frac{1}{\beta_k^{D-1}} \exp{\left(-\frac{d(\mathbf{x}_i, \mathbf{c}_k)}{\beta_k}\right)}
    - w(k, i) \log w(k,i)\;.
\end{equation}
Since the M-step maximises this lower bound only over $\mathbf{c}$ and
$\boldsymbol{\beta}$, the last term becomes a constant and could be omitted
from the optimisation. Further, if all priors $\pi_k$ are assumed constant
$\pi_k = 1/K$, also the first term disappears. What is left is maximisation of
the negative of the loss~\eqref{eq:condensation_loss}.
In the E-step any variational distribution $w(k, i)$ defines a lower bound for
$\mathcal{L}$. It is easy to show that the optimal $w^*(k, i) = p(k|\mathbf{x};
\boldsymbol{\pi}, \boldsymbol{c}, \boldsymbol{\beta}$). Using the Bayes formula
\begin{equation}
    w^*(k, i) = \frac{p(\mathbf{x}_i; \boldsymbol{\pi}, \boldsymbol{c}, \boldsymbol{\beta})}{\sum_{j=1}^K p(\mathbf{x}_i; \boldsymbol{\pi}, \boldsymbol{c}, \boldsymbol{\beta})}
    = \frac{\frac{1}{\beta_k^{D-1}} \exp{\left(-\frac{d(\mathbf{x}_i, \mathbf{c}_k)}{\beta_k}\right)}}{\sum_{j=1}^K \frac{1}{\beta_j^{D-1}} \exp{\left(-\frac{d(\mathbf{x}_i, \mathbf{c}_j)}{\beta_j}\right)}}\;, \label{eq:wopt}
\end{equation}
where the constant a priori probabilities cancel out. Although optimal, when
the embedding dimension $D$ is high (\eg 1024), this formula becomes
numerically unstable. Replacing all $\beta_k$'s by a constant value $\tau$
reduces the formula into~\cref{eq:wtau}. This in principle breaks the EM
monotonicity convergence property, but practically avoids the numerical
instability and allows for independent treatment of the etalon scale parameters
and the soft-to-hard transition parameter $\tau$. In practice, EM converges to
local maxima only even with the optimal lower bound, so we solve both problems
by introducing etalon re-inits.
\subsubsection{Re-Inits.} \label{sec:method_reinits}
K-means, EM algorithm and also the proposed condensation algorithm converges, in general, to a local optimum.
The soft-hard scheduling helps to alleviate this to some extent, but the result still depends on the initialisation. 
Although there exist heuristic initialisation methods like K-means\texttt{++}, they are computationally expensive.
Hence, we introduce etalon re-inits inside the optimisation loop.
For each etalon, we compute its data sample (batch) support as
\begin{equation}
    s(k) = \sum_{i=1}^N w(k, i; \tau)\;.
\end{equation}
In the EM interpretation, this corresponds to a posteriori probability of the
mixture component given data (multiplied by $N$). A small support is indicative
of a misplaced etalon. In the stochastic optimisation, the mini-batch data
samples are often small and biased resulting in
noisy $s(k)$ estimate. Thus, we estimate the support using init-unbiased
running exponential weighted average~\cite{kingma2014adam} over mini-batches.
Given a decay rate $q$, at the iteration $t$ it is estimated as 
\begin{equation}
    s_t(k) = (1 - q_t)s_{t-1}(k) + q_t s(k) \quad \text{where} \quad q_t = q\; / \left(1 - (1-q)^t\right). \label{eq:ewa}
\end{equation}
An etalon is reset if, after a warm-up number of iterations, its $s_t(k) < \theta_r$. The $\theta_r$ is user-defined threshold and it controls the balance between number of used etalons and coverage of low density areas (we set it to $1$ in all experiments).
We reset the etalon to a random data point from the mini-batch and add a little noise to avoid degenerate configurations.

\subsection{Discriminative Classifier} \label{sec:lp}
The discriminative power of simple Linear Probing (LP) technique, used in~\cite{GROOD}, is inadequate for the more complex pixel-level
classification task. 
The LP is generally used to assess zero-shot performance
of large pre-trained model on image classification task and most often is
implemented as a linear regression using off-the-shelf solver (\eg
L-BFGS~\cite{byrd1995}) that requires all data at the same time. 
However, this
techniques are prohibitive in pixel domain due to sheer amount of data. Thus,
we propose to replace the LP by a small multi-layer Perceptron (MLP) network
consisting of two layers with GELU~\cite{hendrycks2023gaussian} non-linearity. That increases the
representation power and is easy to use in the batch processing regime with increased volume of data of the pixel domain.
\subsection{Pixel-Level Adaptations} \label{sec:im2pix}
The pixel-level tasks bring two extra technical challenges: mixed patches and
need for up-scaling. Both are consequence of using ViT~\cite{dosovitskiy2021an}
architecture as a backbone and its first processing step, splitting the input
into patches. Inevitably, some patches contain more than one pixel-level
semantic label (\eg patch on boundary of road and sidewalk will contain labels
of both classes). Although they are relatively rare, we observed their negative
impact on the training of the condensation algorithm. Thus we consider only
patches with more than $90\%$ of one class label during training. At test time,
all patches are processed equally. 
\graynote{Note that the training of the MLP
is performed on the original input resolution by up-scaling the MLP output
logits, which is a common approach, and thus the mixed patches have no effect}.

\pixood as described, works with patches and thus on a lower than
original resolution. To obtain the final full resolution OOD score we follow the
common practice and upscale the output score map. Further, we found out that
up-scaling the encoder's feature maps at the inference time (but \textit{not}
during training) works better than only up-scaling the final OOD scores as
ablated in~\cref{sec:ablation}.

\section{Experiments}
\label{sec:experiments}
In all experiments, the proposed \pixood uses no anomaly or OOD
data during training (such as auxiliary datasets or
synthetic anomalies through augmentation) except for the LaRS dataset where the
obstacle class is part of training data. We use frozen
DINOv2~\cite{oquab2023dinov2} ViT-L variant as a backbone in all experiments
and train using AdamW optimizer: the condensation algorithm (100 epochs, starting learning rate $0.1$ with cosine decay and `budget' $K$ set to
$1000$), classification MLP (30 epochs with learning rate $0.0001$). The final calibrated strategy is found on available training
data for respective task. Qualitative results with typical failure cases are
shown in~\cref{fig:intro,fig:failure_cases} and more examples are provided in
supplementary material. All training was done on a single NVIDIA A100-40GB. The runtime of
the condensation algorithm for CityScapes dataset and for all classes was 5, 8 and 17 hours for K equal to 200, 500 and 1000 respectively.

{\bf Datasets.} Commonly used benchmarks from recent
literature for the respective domains are used: (i) for the road anomaly detection, where the evaluation is limited to the road region only -- Road
Anomaly~\cite{Lis_2019_ICCV}, FischyScapes LaF~\cite{pinggera2016lost,
blum2019fishyscapes}, SMIYC~\cite{chan_smiyc_2021} (Obstacle Track, LaF
NoKnown), (ii) for semantic segmentation anomaly  -- SMIYC (Anomaly Track), 
(iii) for industrial inspection -- MVTec AD~\cite{Bergmann_2019_CVPR} and 
(iv) for maritime obstacle detection -- LaRS~\cite{Zust2023LaRS}.

{\bf Evaluation metrics.} We use metrics standard for the respective tasks. For
road and semantic segmentation anomaly detection, the commonly used metrics are
Average Precision (AP) -- area under precision-recall curve, False Positive
Rate at $95\%$ of True Positive Rate, denoted simply as FPR, and mean F1 score computed
component-wise and averaged over different detection thresholds. In the industrial
inspection domain, image and pixel level AUROC -- area under ROCs curve are most
often used together with AUPRO used in~\cite{Bergmann_2019_CVPR} -- per-region
overlap (PRO), to account for anomalies of different sizes within one image. The
maritime obstacle detection (LaRS) uses standard mean Intersection over Union (mIoU)
for the semantic classes and a domain specific metrics~\cite{Bovcon2021} such
as: 1) wateredge accuracy -- computed from boundary between water and
static obstacles and (2) obstacle detection accuracy -- precision, recall
and F1 score where the true positive detections are considered if predicted
obstacle covers the ground-truth pixels from more than $70\%$. The final
ranking in LaRS is determined by the Q measure which is computed as
$\text{F1}\times\text{mIoU}$.
\begin{table*}[t]
\parbox{.48\linewidth}{
    \caption{Ablation study -- novel condensation, ``budget'' $K$, and
       pixel adaptation techniques.
       All reported results use the best performing setup of the \pixood pipeline except one change that is being ablated. 
       Components of the best setup are highlighted in green.}
    \centering\resizebox{0.48\textwidth}{!}{
       \begin{tabular}{lrcc}
       \toprule
       \multirow{2}{*}{\pixood with}& \multirow{2}{*}{$K$} & \multicolumn{2}{c}{RA+FS+RO} \\
       \cmidrule(lr){3-4}
       {} & {} & mean AP $\uparrow$ & mean FPR $\downarrow$ \\
       \midrule
       Single Mean                                                 & 1                  & 88.14 & 3.49 \\ 
       Soft K-Means -- \cref{eq:skmeans_tau}                       & 200                & 93.01 & 2.26 \\ 
       Soft K-Medians -- \cref{eq:skmedians}                       & 200                & 93.54 & 1.99 \\ 
       \cmidrule(lr){1-4}
       \selectedcell Soft-to-Hard Cond. -- \cref{sec:condensation} & 200                & 94.11 & 1.82 \\ 
       \phantom{Soft-to-Hard Cond.}                                & 500                & 94.71 & 1.67 \\ 
       \phantom{Soft-to-Hard Cond.}                                & \selectedcell 1000 & 94.80 & 1.64 \\ 
       \cmidrule(lr){1-4}
       Linear Probe                                                & {}                 & 92.36 & 2.38 \\ 
       \selectedcell Multi-Layer Perceptron                        & {}                 & 94.80 & 1.64 \\ 
       \cmidrule(lr){1-4}
       Patch label percentage $>$ 50\%                             & {}                 & 94.22 & 1.77 \\ 
       \selectedcell Patch label percentage $>$ 90\%               & {}                 & 94.80 & 1.64 \\ 
       \cmidrule(lr){1-4}
       Feature map up-sampling $\times$ 1                          & {}                 & 92.14 & 2.35 \\ 
       Feature map up-sampling $\times$ 2                          & {}                 & 94.20 & 1.75 \\ 
       Feature map up-sampling $\times$ 5                          & {}                 & 94.78 & 1.65 \\ 
       \selectedcell Feature map up-sampling $\times$ 7            & {}                 & 94.80 & 1.64 \\ 
       \bottomrule
       \end{tabular}
    }
    \label{tab:fsra_ablation_k}
}
\hfill
\parbox{.48\linewidth}{
    \caption{Comparison with state-of-the-art methods on Road Anomaly and FS LaF datasets.
    The gray-out lines show additional results of the DaCUP method
    with DINOv2 backbone and a \textit{meta}-method obtained by combining
    \pixood with road-anomaly specialised method DaCUP~\cite{Vojir_2023_WACV}
    (for details see~\cref{sec:exp_road_anomaly}).}

    \centering\resizebox{0.48\textwidth}{!}{
    \begin{tabular}{lcrrrr}
    \toprule
    \multirow{2}{*}{Method} & \multirow{2}{*}{\shortstack[c]{OOD\\ Data}} & \multicolumn{2}{c}{Road Anomaly} & \multicolumn{2}{c}{FishyScapes LaF} \\
    \cmidrule(lr){3-4}
    \cmidrule(lr){5-6}
    & & AP $\uparrow$ & FPR $\downarrow$ & AP $\uparrow$ & FPR $\downarrow$ \\
    \midrule
    SynBoost~\cite{Di_Biase_2021_CVPR}     & \cmark & 63.72         & 52.27        & 92.46         & \second 0.66 \\
    PEBAL~\cite{Tian2021PixelwiseEA}       & \cmark & 67.31         & 39.41        & 84.05         & 1.58 \\
    Maximized Entropy                      & \cmark & \second 96.23 & \second 6.03 & 77.16         & 10.14 \\
    RbA~\cite{nayal2023ICCV}               & \cmark & 92.99         & 9.01         & 85.98         & 2.87 \\
    Mask2Anomaly~\cite{Rai_2023_ICCV}      & \cmark & 94.10         & 16.91        & \first 94.49  & \first 0.52 \\
    EAM~\cite{Grcic_2023_CVPR}             & \cmark & \first 96.43  & \first 2.26  & \second 93.67 & 0.77 \\
    \cmidrule(lr){1-6}
    Image Resynthesis~\cite{Lis_2019_ICCV} & \xmark & 76.39         & 48.08        & 66.75         & 3.10 \\
    EAM~\cite{Grcic_2023_CVPR}             & \xmark & 88.96         & 17.89        & 64.86         & 16.73 \\
    RbA~\cite{nayal2023ICCV}               & \xmark & 89.76         & 16.54        & 74.70         & 15.13 \\
    JSR-Net~\cite{Vojir_2021_ICCV}         & \xmark & 94.42         & 9.25         & 78.30         & 3.96 \\
    DaCUP~\cite{Vojir_2023_WACV}           & \xmark & \second 96.19 & \second 5.46 & \second 89.75 & \second 1.45 \\
    \pixood ({\bf Ours})                   & \xmark & \first 96.39  & \first 4.30  & \first 93.55  & \first 0.54 \\
    \cmidrule(lr){1-6}
    \rowcolor{gray!10}
    \DaCUPpp (CityScapes)                  & \xmark & 98.24         & 3.04         & 93.23         & 0.57\\
    \rowcolor{gray!10}
    Ours\texttt{+}DaCUP (recon.)           & \xmark & 98.66         & 2.12         & 94.66         & 0.30 \\
    \bottomrule
    \end{tabular}
    }
    \label{tab:rafs_results}
}
\end{table*}%

%
\subsection{Ablation Study} \label{sec:ablation}
We use road anomaly detection as the downstream task on which we
demonstrate the effectiveness of the proposed condensation algorithm and the effect
of different `budget' size $K$.
All hyper-parameters and random seeds for different variants
of the method were fixed.

The results for both these ablations are presented
in~\cref{tab:fsra_ablation_k}.
They show that a single mean representation is
insufficient for complex class modelling in pixel-level domain. Furthermore,
robustness and efficacy of the proposed condensation method is demonstrated by
increased performance compared to K-\{Means, Medians\} algorithms. The
performance can be further improved by allowing larger `budget' $K$ and it
saturates for $K$ around $1000$ suggesting that the proposed method can utilise extra
clusters, up to a saturation point, efficiently.

Furthermore, we evaluate the proposed generalisations and technical
improvements from~\cref{sec:im2pix}. 
\graynote{Note that for \textit{Linear Probe}
we used single fully connected layer and trained it by SGD, since it can be
processed in batches as discussed in~\cref{sec:lp}}.
Results are presented in a leave-one-out manner in the bottom three blocks of~\cref{tab:fsra_ablation_k}. By modifying only one part
and leaving the rest in the best configuration it clearly shows the benefit of every design choice.
\begin{figure*}[t]
  \centering
  \includegraphics[width=0.95\textwidth]{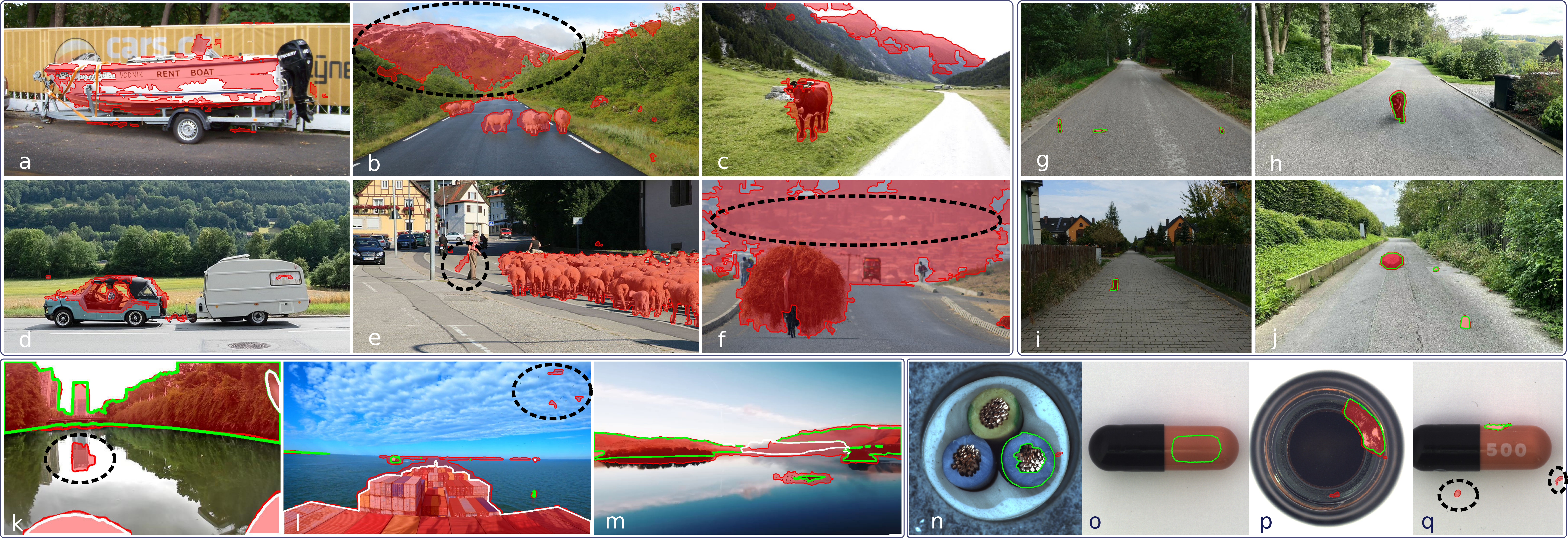}
  \caption{Typical outputs with focus on "failure" cases. \pixood anomalies are
    sometimes under-segmented (a) or over-segmented (b,c,m), but it often finds
    unexpected but reasonable anomalies: a stick (e), reflection in the water
    (k), birds (l), extra scratches (p), pill remains (q). It is also unable
    to detect logical anomalies like the switched cable in (n) or missing label
    in (o). Moreover, semantic/domain shifts are considered as anomaly: mountains
    (b,c), convertible roof with see-through (d) and city view (f) which are
    not present in the CityScapes. Legend: red - detected anomaly, green - GT when available,
    white - ignore region; ellipses mark relevant
    regions.\label{fig:failure_cases}}
\end{figure*}
\subsection{Road Anomaly Detection}
\label{sec:exp_road_anomaly}
We compare the proposed method with recent
state-of-the-art methods on two standard datasets, Road Anomaly and FishyScapes
LaF, and also on SMIYC benchmark designed for road anomaly segmentation task. 
Results  on the standard datasets are presented and compared with most recent publications~\cite{nayal2023ICCV, Grcic_2023_CVPR,Rai_2023_ICCV, Vojir_2023_WACV}
in~\cref{tab:rafs_results}. 
The proposed method outperforms other state-of-the-art methods (even several of those that
utilise an auxiliary dataset to model the OOD data) often by large margin.

Evaluation results on the SMIYC benchmark are reported in~\cref{tab:smiyc_results_main}. We
report all three tracks, two for road anomaly detection (Obstacle Track and LaF
NoKnown) and one for semantic segmentation anomaly detection (Anomaly Track).
Compared to methods not utilising any auxiliary OOD data, our method is the
best performing in Obstacle Track and second best in LaF NoKnown
track. 

To investigate the effect of a strong pre-trained representation (even though
an extensive evaluation is out of the scope of this paper), we enhanced the
DaCUP~\cite{Vojir_2023_WACV} method with the modern
DINOv2~\cite{oquab2023dinov2} ViT-L backbone. We denote the method as \DaCUPpp.
In \DaCUPpp we replace the DeepLabv3 segmentation part by the \pixood
classification MLP with the weights fixed and train the rest of DaCUP on
CityScapes as in~\cite{Vojir_2023_WACV}. \DaCUPpp improves most notably by
decreasing the false positive rate in all experiments. The \pixood remains 
competitive and outperforms the \DaCUPpp on challenging scenarios of SMIYC.

Furthermore, since our method is generic (making no task specific design choices), it is of
interest to try to incorporate DaCUP as a specialised module for
a particular task. We added the DaCUP Reconstruction Module specialised for road anomaly
detection as a second OOD detection head into \pixood. The
final anomaly score is a weighted average of \pixood score ($75\%$) and DaCUP module road anomaly
score ($25\%$). The resulting task-augmented
\pixood (denoted as ``Ours\texttt{+}DaCUP (recon.)'') significantly boosts the results
of the \pixood, outperforms \DaCUPpp and shows that domain specific knowledge and generic OOD detection approach are
complementary to a degree.
\begin{table*}[t]
\caption{Comparison with the state-of-the-art on the SMIYC~\cite{chan_smiyc_2021}
    benchmark. For clarity, only top performing methods and main
    evaluation metrics are shown.
    The gray-out lines show the same comparison as in~\cref{tab:rafs_results} (for details see~\cref{sec:exp_road_anomaly}).}
\centering\resizebox{0.95\textwidth}{!}{
\begin{tabular}{lcrrrrrrrrr}
\toprule

\multirow{2}{*}{Method} & \multirow{2}{*}{\shortstack[c]{OOD\\ Data}} & \multicolumn{3}{c}{Anomaly Track} & \multicolumn{3}{c}{Obstacle Track} & \multicolumn{3}{c}{LaF NoKnown}\\

\cmidrule(lr){3-5}
\cmidrule(lr){6-8}
\cmidrule(lr){9-11}
& & 
\multicolumn{1}{c}{AP $\uparrow$}& \multicolumn{1}{c}{FPR $\downarrow$}&\multicolumn{1}{c}{mean F1 $\uparrow$}&
\multicolumn{1}{c}{AP $\uparrow$}& \multicolumn{1}{c}{FPR $\downarrow$}&\multicolumn{1}{c}{mean F1 $\uparrow$}&
\multicolumn{1}{c}{AP $\uparrow$}& \multicolumn{1}{c}{FPR $\downarrow$}&\multicolumn{1}{c}{mean F1 $\uparrow$}\\

\midrule

SynBoost~\cite{Di_Biase_2021_CVPR}      & \cmark & 56.44         & 61.86 & 9.99  & 71.34 & 3.15  & 37.57 &\first 81.71 &\second 4.64  & 48.72\\
ATTA~\cite{gao2023atta}                 & \cmark & 67.04         & 31.57 & 20.64 & 76.46 & 2.81  & 36.57 & --    & --    & --   \\
DenseHybrid~\cite{grcic22eccv}          & \cmark & 77.96         & 9.81  & 31.08 & 87.08 & 0.24  & 50.72 &\second 78.67 &\first 2.12  &\first 52.33\\
RPL\texttt{+}CoroCL~\cite{Liu_2023_ICCV}         & \cmark & 83.49         & 11.68 & 30.16 & 85.93 & 0.58  & 56.69 & --    & --    & --   \\
Maximized Entropy~\cite{Chan_2021_ICCV} & \cmark & 85.47         & 15.00 & 28.72 & 85.07 & 0.75  & 48.51 & 77.90 & 9.70  &\second 49.92\\
Mask2Anomaly~\cite{Rai_2023_ICCV}       & \cmark & 88.72         & 14.63 & 47.16 &\second 93.22 &\second 0.20  &\second 68.15 & --    & --    & --   \\
EAM~\cite{Grcic_2023_CVPR}              & \cmark & \second 93.75 &\first 4.09  &\first 60.86 & 92.87 & 0.52  &\first 75.58 & --    & --    & --   \\
RbA~\cite{nayal2023ICCV}                & \cmark & \first 94.46  &\second 4.60  &\second 51.87 &\first 95.12 &\first 0.08  & 57.44 & --    & --    & --   \\

\cmidrule(lr){1-11}

ODIN~\cite{liang2018enhancing}          & \xmark & 33.06         & 71.68 & 5.15  & 22.12 & 15.28 & 9.37  & 52.93 & 30.04 & 34.53\\
JSRNet~\cite{Vojir_2021_ICCV}           & \xmark & 33.64         & 43.85 & 13.66 & 28.09 & 28.86 & 11.02 & 74.17 & 6.59  & 35.97\\
PEBAL~\cite{Tian2021PixelwiseEA}        & \xmark & 49.14         & 40.82 & 14.48 & 4.98  & 12.68 & 5.54  & --    & --    & --   \\
Image Resynthesis~\cite{Lis_2019_ICCV}  & \xmark & 52.28         & 25.93 & 12.51 & 37.71 & 4.70  & 8.38  & 57.08 & 8.82  & 19.17\\
Road Inpainting~\cite{lis2023detecting} & \xmark & --            & --    & --    & 54.14 & 47.12 & 36.01 & 82.93 & 35.75 &\second 52.25\\
NFlowJS~\cite{grcic2023dense}           & \xmark & 56.92         & 34.71 & 14.89 & 85.55 &\second 0.41  & 50.36 &\first 89.28 &\first 0.65  &\first 61.75\\
ObsNet~\cite{besnier2021trigger}        & \xmark & 75.44         & 26.69 &\second 45.08 & --    & --    & --    & --    & --    & --   \\
DaCUP~\cite{Vojir_2023_WACV}            & \xmark & --            & --    & --    & 81.50 & 1.13  & 46.01 & 81.37 & 7.36  & 51.14\\
RbA~\cite{nayal2023ICCV}                & \xmark & \second 86.13 & 15.94 & 42.04 &\second 87.85 & 3.33  & 50.42 & --    & --    & --   \\
CSL~\cite{Zhang_Li_Qi_Yang_Ahuja_2024}  & \xmark & 80.08         & \first \phantom{1}7.16  & \first 50.39 & 87.10 & 0.67 & \first 51.02 & -- & -- & -- \\
cDNP~\cite{Galesso_2023_ICCV}           & \xmark & \first 88.90  & \second 11.42 & 28.12 & --    & --    & --    & --    & --    & --   \\

\pixood ({\bf Ours})                    & \xmark & 68.88         & 54.33 & 19.82 &\first 88.90 &\first 0.30  & \second 50.82 &\second 85.07 &\second 4.46  & 44.41\\

\cmidrule(lr){1-11}

\rowcolor{gray!10}
\DaCUPpp (CityScapes)                    & \xmark & -- & -- & -- & 86.10 & 0.90 & 44.61 & 79.47 & 3.40 & 37.60 \\

\rowcolor{gray!10}
Ours\texttt{+}DaCUP (recon.)                   & \xmark & -- & -- & -- & 84.96 & 0.93 & 39.44 & 87.57 & 1.29 & 48.97 \\

\bottomrule
\end{tabular}
}
\label{tab:smiyc_results_main}
\end{table*}%

Anomaly Track is the only experiment where \pixood is ''formally''
under-performing; The typical observed failures are: (i) under-segmenting, (ii)
detecting semantic shifts, such as mountains or forests which are nor annotated in
Cityscapes dataset and (iii) detecting domain shift, \ie same class as in
Cityscapes but with very different appearance. The semantic/domain shifts may not be considered OOD
in the Anomaly Track due to (vague) definition of anomaly and it requires an
algorithm to generalize semantic concepts (like a car). Instead, our method
detects anomalies w.r.t. the given ID data. This allows detecting these semantic/domain shifts,
which is important in many tasks. For qualitative examples
see~\cref{fig:failure_cases}.
\begin{table*}[t]
\parbox{.48\linewidth}{
    \caption{Comparison with the state-of-the-art on MVTec AD~\cite{Bergmann_2019_CVPR}. The methods are grouped by the use of a per-class models for each object category vs unified model.
    The bottom line shows results of combining \pixood with specialised method DRAEM~\cite{Zavrtanik_2021_ICCV}.}
    \centering\resizebox{0.48\textwidth}{!}{
    \begin{tabular}{lccrrr}
    \toprule
    \multirow{2}{*}{Method} & \multirow{2}{*}{\shortstack[c]{per-object\\ model}} & \multirow{2}{*}{\shortstack[c]{OOD\\ Data}} & Image-level & \multicolumn{2}{c}{Pixel-level} \\

    \cmidrule(lr){4-4}
    \cmidrule(lr){5-6}

    {} & {} & {} & AUROC $\uparrow$ & AUPRO $\uparrow$ & AUROC $\uparrow$ \\
        
    \midrule

    PaDiM~\cite{defard2021}                & \cmark & \xmark & 97.9        \phantom{00}& 92.1         \phantom{00}& 97.5 \phantom{00}\\ 
    DRAEM~\cite{Zavrtanik_2021_ICCV}       & \cmark & \cmark & 98.0        \phantom{00}& 92.8         \phantom{00}& 97.3 \phantom{00}\\ 
    DAF~\cite{cai2023}                     & \cmark & \cmark & 97.6        \phantom{00}& 93.0         \phantom{00}& \second 98.1 \phantom{00}\\ 
    CFLOW-AD~\cite{Gudovskiy_2022_WACV}    & \cmark & \xmark & 98.3        \phantom{00}& \first 98.6  \phantom{00}& 94.6 \phantom{00}\\ 
    PatchCore (25\%)~\cite{Roth_2022_CVPR} & \cmark & \xmark & \second 99.1 \phantom{00}& 93.4         \phantom{00}& \second 98.1 \phantom{00}\\ 
    PNI~\cite{Bae_2023_ICCV}               & \cmark & \xmark & \first 99.6 \phantom{00}& \second 96.1 \phantom{00}& \first 99.0 \phantom{00}\\ 
    \cmidrule(lr){1-6}
    UniAD~\cite{you2022a}                  & \xmark & \xmark & 96.5 \phantom{00}& --           \phantom{00}& \second 96.8 \phantom{00}\\ 
    HVQ-Trans~\cite{lu2023hierarchical}    & \xmark & \xmark & \first 98.0 \phantom{00}& --           \phantom{00}& \first 97.3 \phantom{00}\\ 
    \pixood ({\bf Ours})                         & \xmark & \xmark & \second 97.0 \phantom{00}& 91.3         \phantom{00}& 94.6 \phantom{00}\\ 

    \cmidrule(lr){1-6}
    \rowcolor{gray!10}
    Ours\texttt{+}DRAEM                           & \cmark & \cmark & 98.6     \phantom{00}& 95.3      \phantom{00}& 98.0 \phantom{00}\\

    \bottomrule
    \end{tabular}
    }%
    \label{tab:mvtec_results}
}
\hfill
\parbox{.48\linewidth}{
    \caption{Comparison with the state-of-the-art on LaRS~\cite{Zust2023LaRS}.
     Results from official leaderboard (\url{https://lojzezust.github.io/lars-dataset/}) 
     at the date of submission (considering latest model submissions from 
     individual institutions; UniL - University of Ljubljana, UniC - University of Cagliary).} 
    \centering\resizebox{0.48\textwidth}{!}{
    \hspace{-1ex}
    \begin{tabular}{lcccccc}
    \toprule
    Model Name & \textbf{Q} & $\mu$ & Pr & Re & F1 & mIoU \\

    \midrule

    DeepLabv3 (UniL) & {\color{gray!15}\ding{189}} 62.9 & 77.5 & 61.1 & 72.0 & 66.1 & 95.2\\
    SegFormer (UniL) & {\color{gray!30}\ding{188}} 67.8 & 78.6 & 63.8 & 77.5 & 70.0 & 96.8\\
    Swin-T (UniL)    & {\color{gray!45}\ding{187}} 71.3 & 78.8 & 67.6 & 80.4 & 73.4 & 97.2\\
    Mask2Former (HSU)             & {\color{gray!60}\ding{186}} 73.8 & 78.5 & 76.9 & 73.8 & 75.3 & 98.0\\
    Mask2Former (DLMU)            & {\color{gray!90}\ding{184}} 75.7 & 78.4 & 79.7 & 75.1 & 77.3 & 97.8\\
    Swin-L (HKUST)                & \second 77.8                     & 79.6 & 78.5 & 82.0 & 80.2 & 97.1\\
    Snarciv3 (UniC)   & \first 78.1                      & 79.7 & 76.9 & 83.0 & 79.9 & 97.8\\
    \pixood ({\bf Ours})                & {\color{gray!75}\ding{185}} 73.9 & 74.6 & 70.7 & 81.6 & 75.8 & 97.5\\
    \bottomrule
    \end{tabular}
    }
    \label{tab:lars_results}
}
\end{table*}
\subsection{Industrial Anomaly Detection} \label{sec:exp_industrial_anomaly}

The industrial anomaly detection problem is most often addressed by training
a model for each individual class (product). Only recently, new methods started to address the
problem in a holistic manner by training a single model for all objects categories. 
\pixood inherently falls into this category of a single unified model. The results for the standard MVTec AD~\cite{Bergmann_2019_CVPR}
benchmark are shown in~\cref{tab:mvtec_results} and per category performance in the supplementary materials.
\pixood performs competitively to domain specialised method that use
unified model. 

Similarly to the road anomaly task, we can incorporate DRAEM~\cite{Zavrtanik_2021_ICCV} method domain specific knowledge to increase the performance for this specific task. 
The resulting task-augmented \pixood (denoted as ``Ours\texttt{+}DRAEM'')
improves the results (significantly in case of AUPRO metric).

\subsection{Maritime Obstacle Detection} \label{sec:exp_maritime_obstacle}

This experiment demonstrates the versatility of the proposed method in 
different settings and domains. The maritime obstacle detection problem is
posed as semantic segmentation. The task is to segment the image into
three classes -- {\it Water, Sky and Obstacle}. During training, examples of
all classes are available. To fit \pixood to the setting, we model all
three classes as in-distribution data and during the inference we use the
classification network to produce class labelling. For each pixel and predicted
class we then decide based on the OOD score if the pixel is ID (the
class label remains as predicted) or OOD for which we change the class label to
{\it Obstacle}. The results of our method are compared to
stat-of-the-art methods (taking the latest version of the method submitted)
from the official leaderboard and shown in \cref{tab:lars_results}. The
proposed method performs favourably placing fourth place with the main issue
being low precision due to over-segmentation (see~\cref{fig:failure_cases} for
qualitative examples).

\section{Conclusions} \label{sec:conclusions}

In this paper we propose a novel pixel-level OOD detection method. The method
is not designed for a specific task or benchmark, and it performs competitively
on a range of pixel-level OOD problems. The method requires no OOD training
samples, neither real nor synthetic. The method builds on a proposed data
condensation algorithm which is theoretically linked to the
optimisation of a complete data log likelihood in the EM algorithm.
We applied the proposed method to three very diverse pixel-level anomaly
benchmarks and achieved state-of-the-art results on four out of seven
considered datasets.

\noindent{\bf Acknowledgement}
This work was supported by Toyota Motor Europe and Czech Technical University in Prague institutional support Future Fund.

\bibliographystyle{splncs04}
\bibliography{main}

\clearpage
\appendix 
\section{Social, Ethics and Privacy Impact and Considerations}
We thank anonymous {\bf Reviewer \verb|#|1} for bringing these concerns to our attention:
\begin{quote}
\it
The authors state that there is no potential negative societal impact in their
research. However, thinking about a real-world deployment of the model proposed
for, e.g., anomaly detection in video surveillance scenarios, I think it would
be nice to briefly discuss ethical, privacy, transparency and fairness concerns
in this paper.
\end{quote}
In response, we considered and discussed the issue broadly. No immediate
privacy or ethical issues with the current pixel-level method were identified.
On the contrary, we argue that enhancing deployed models with the ability to
output the "I do not know" label is beneficial for AI systems in general.
Representing decision uncertainty partially mitigates the lack of diversity in
datasets, potential biases, and it avoids non-confident or wrong over-confident
decisions in downstream tasks by identifying these cases as out-of-distribution (OOD). 

We, however, agree that in a future research there may arise privacy and
ethical concerns regarding applications in domains such as face
re-identification and recognition, surveillance, or behavior and action
analysis. It is possible that a method, similar to the proposed \pixood, could
potentially be exploited for nefarious applications. For example, penalizing
``not approved'' (thus not in the training data and  identified as OOD) types
of behavior or actions performed  under surveillance. The proposed method is
currently not  directly applicable in such settings and the adaptation, in our
expert opinion, is non-trivial. 

We keep these risks in mind, noting that every tool can be used in both
socially beneficial or harmful ways.

\clearpage
\section{Additional Quantitative Results}
\begin{wraptable}{r}{0.47\textwidth}
\vspace{-3.5em}
\caption{Per-object results of the proposed \pixood method on the MVTec AD~\cite{Bergmann_2019_CVPR} benchmark.}
\centering\resizebox{0.42\textwidth}{!}{
\begin{tabular}{clccc}
\toprule
{} & \multirow{2}{*}{Class} & Image-level & \multicolumn{2}{c}{Pixel-level} \\
\cmidrule(lr){3-3}
\cmidrule(lr){4-5}
{} & {} & AUROC $\uparrow$ & AUPRO $\uparrow$ & AUROC $\uparrow$ \\

\midrule
\multirow{10}{*}{\rotatebox[origin=c]{90}{Objects}} & bottle     & 100.00 & 96.65 & 98.84\\
{}                                                  & cable      & 93.39  & 90.05 & 95.51\\
{}                                                  & capsule    & 94.815 & 95.83 & 97.36\\
{}                                                  & hazelnut   & 99.89  & 98.16 & 99.40\\
{}                                                  & metal nut  & 100.00 & 92.62 & 96.64\\
{}                                                  & pill       & 96.37  & 95.15 & 87.90\\
{}                                                  & screw      & 83.05  & 87.20 & 95.99\\
{}                                                  & toothbrush & 92.5   & 66.94 & 73.83\\
{}                                                  & transistor & 97.33  & 75.61 & 89.75\\
{}                                                  & zipper     & 98.77  & 87.46 & 94.96\\
\cmidrule(lr){1-5}
\multirow{5}{*}{\rotatebox[origin=c]{90}{Textures}} & carper     & 99.40  & 96.80 & 99.01\\
{}                                                  & grid       & 100.00 & 98.27 & 99.61\\
{}                                                  & leather    & 100.00 & 99.13 & 99.40\\
{}                                                  & tile       & 100.00 & 93.87 & 97.21\\
{}                                                  & wood       & 99.47  & 95.23 & 93.16\\
\bottomrule
\end{tabular}
}
\label{tab:mvtecad_per_class_supp}
\vspace{-8em}
\end{wraptable}

Additional results on the SMIYC benchmark for all tracks and all metrics are reported in~\cref{tab:smiyc_results_all} and 
per-class results of the proposed \pixood method on the MVTec AD benchmark are reported in~\cref{tab:mvtecad_per_class_supp}.

\section{Additional Qualitative Results}
\begin{figure*}[t]
  \centering
    \includegraphics[width=0.95\textwidth]{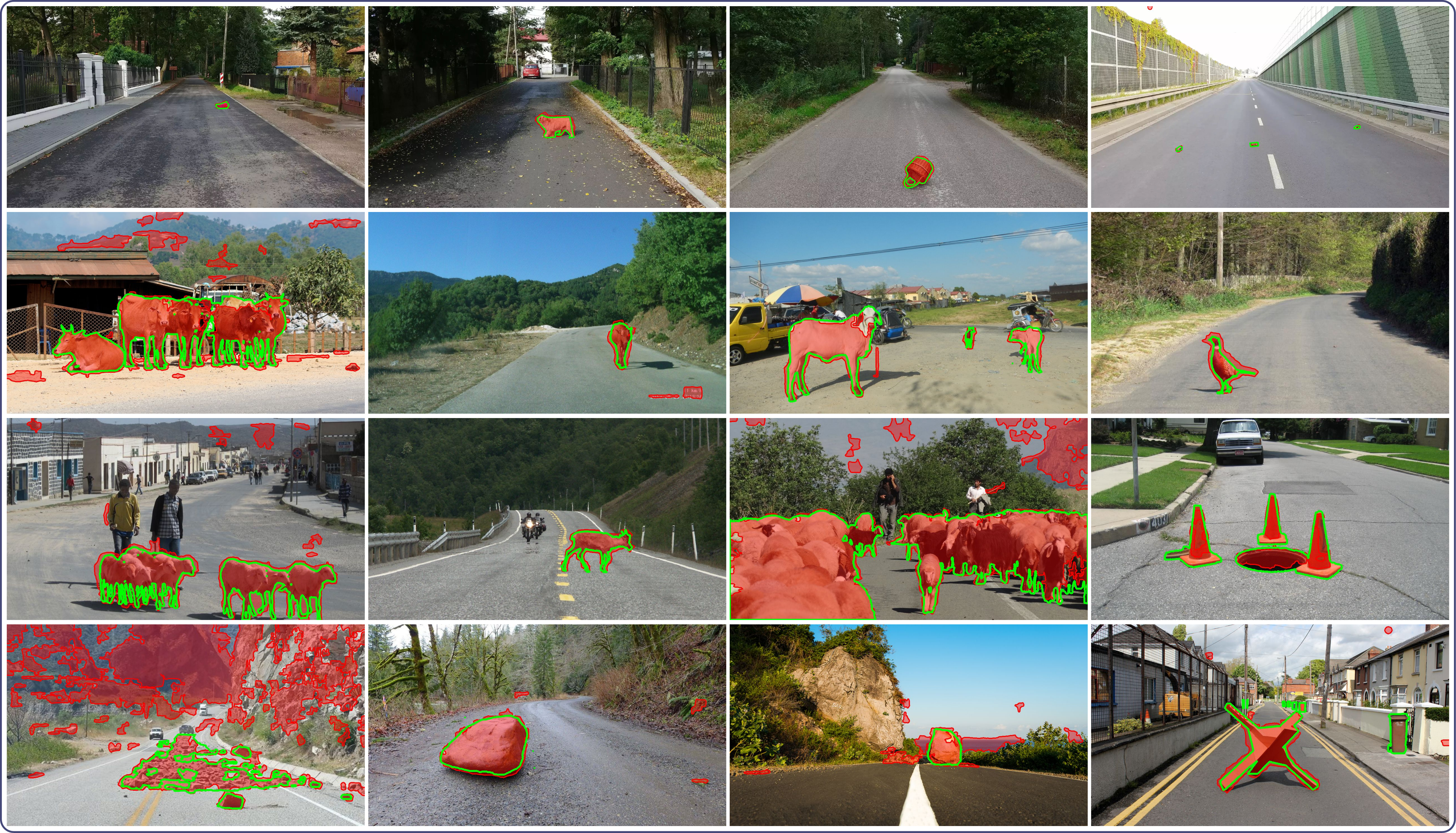}
  \caption{Road anomaly detection examples.}
  \label{fig:roadanomaly_supp}
\end{figure*}

\begin{figure*}[t]
  \centering
    \includegraphics[width=0.95\textwidth]{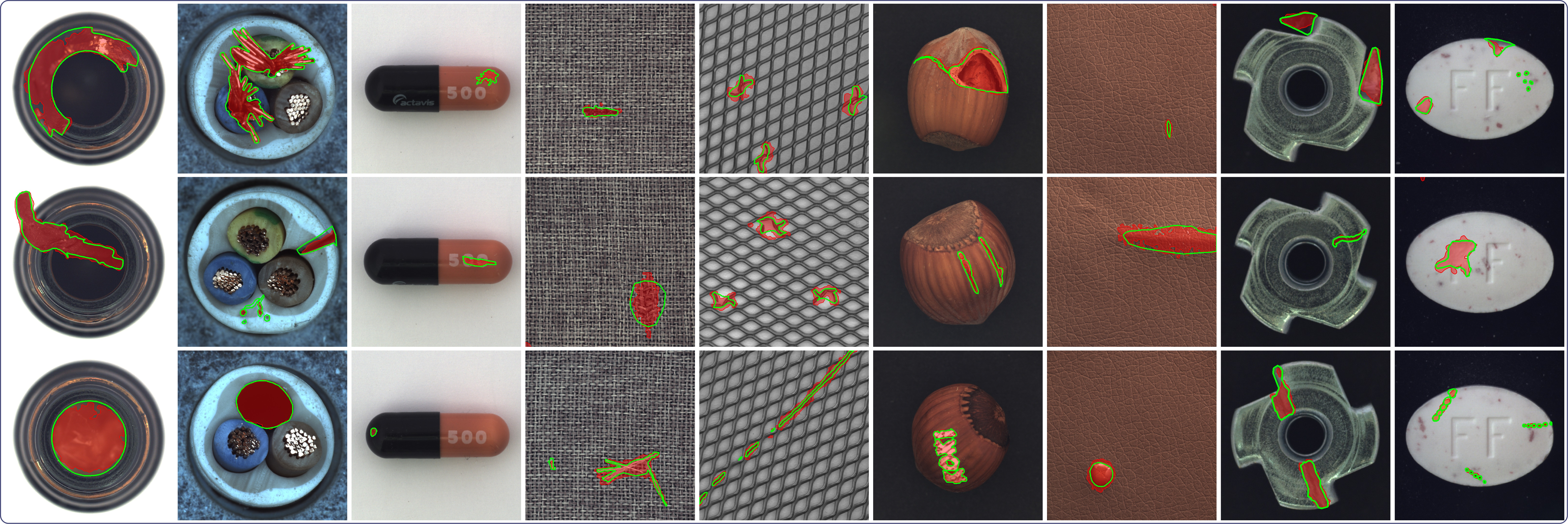}
  \caption{Industrial MVTec AD detection examples.}
  \label{fig:industrial_supp}
\end{figure*}

\begin{figure*}[t]
  \centering
    \includegraphics[width=0.95\textwidth]{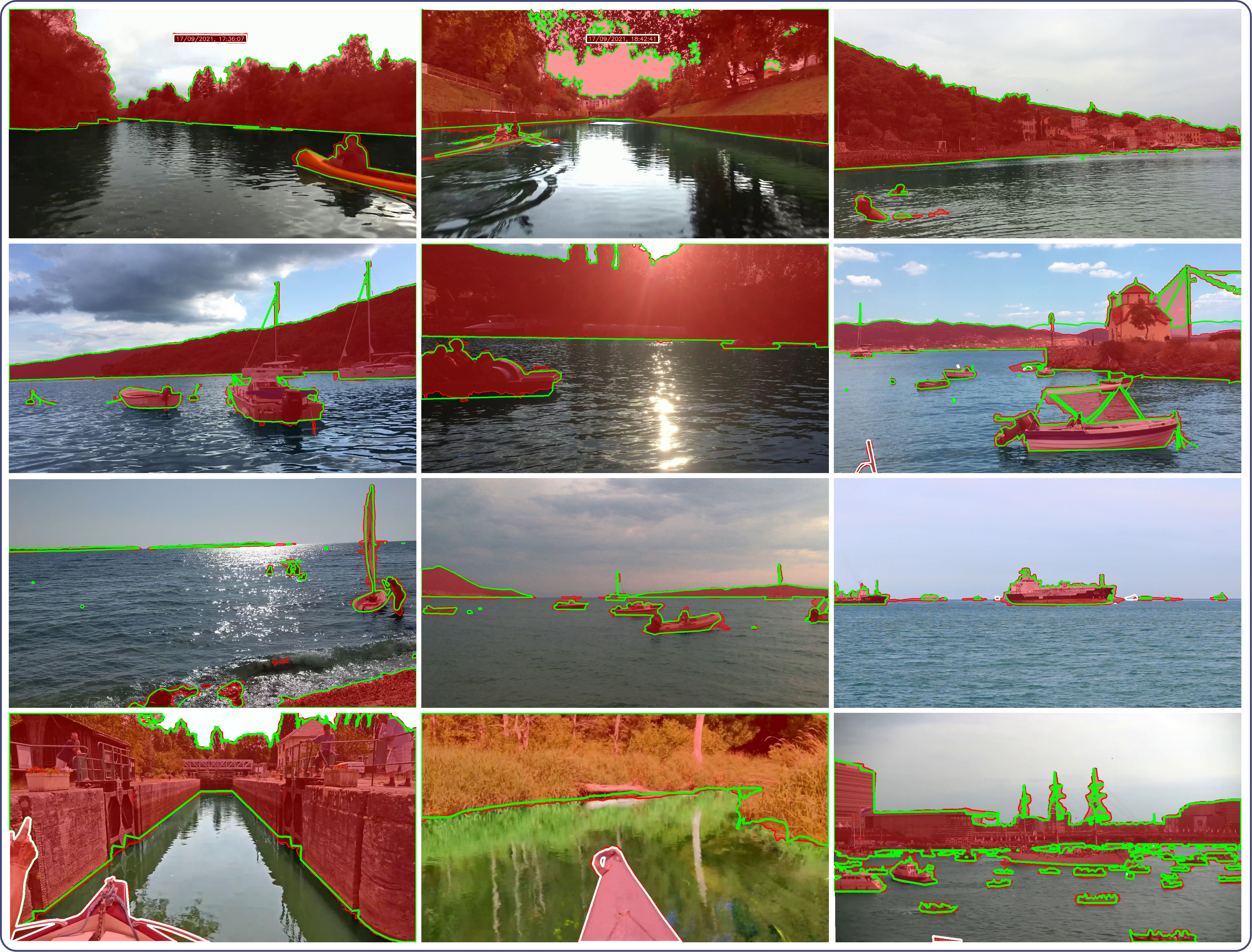}
    \caption{Maritime Obstacle (LaRS) anomaly detection examples.}
  \label{fig:maritime_supp}
\end{figure*}

We present additional qualitative results for road
anomaly detection in~\cref{fig:roadanomaly_supp}, industrial anomaly detection in~\cref{fig:industrial_supp} and maritime
obstacle detection in~\cref{fig:maritime_supp}. 
\begin{table*}[h]
\caption{Comparison with the state-of-the-art on SMIYC~\cite{chan_smiyc_2021} benchmark.}
\centering\resizebox{0.95\textwidth}{!}{
\begin{tabular}{lcccccccccccccccccc}
\toprule
\multirow{2}{*}{Method} & \multirow{2}{*}{\shortstack[c]{OOD\\ Data}} & \multicolumn{5}{c}{Anomaly Track} & \multicolumn{5}{c}{Obstacle Track} & \multicolumn{5}{c}{LostAndFound NoKnown} \\
\cmidrule(lr){3-7}
\cmidrule(lr){8-12}
\cmidrule(lr){13-17}
& & 
AP $\uparrow$& FPR $\downarrow$&sIoU gt $\uparrow$& PPV $\uparrow$&mean F1 $\uparrow$& 
AP $\uparrow$& FPR $\downarrow$&sIoU gt $\uparrow$& PPV $\uparrow$&mean F1 $\uparrow$& 
AP $\uparrow$& FPR $\downarrow$&sIoU gt $\uparrow$& PPV $\uparrow$&mean F1 $\uparrow$\\ 
\midrule

SynBoost~\cite{Di_Biase_2021_CVPR}          & \cmark & 56.44 & 61.86 & 34.68 & 17.81 & 9.99  & 71.34 & 3.15  & 44.28 & 41.75 & 37.57 & 81.71 & 4.64  & 36.83 & 72.32 & 48.72\\
ATTA~\cite{gao2023atta}              & \cmark & 67.04 & 31.57 & 44.58 & 29.55 & 20.64 & 76.46 & 2.81  & 43.93 & 37.66 & 36.57 & --    & --    & --    & --    & --   \\
DenseHybrid~\cite{grcic22eccv}       & \cmark & 77.96 & 9.81  & 54.17 & 24.13 & 31.08 & 87.08 & 0.24  & 45.74 & 50.10 & 50.72 & 78.67 & 2.12  & 46.90 & 52.14 & 52.33\\
RPL+CoroCL~\cite{Liu_2023_ICCV}        & \cmark & 83.49 & 11.68 & 49.77 & 29.96 & 30.16 & 85.93 & 0.58  & 52.62 & 56.65 & 56.69 & --    & --    & --    & --    & --   \\
Maximized Entropy~\cite{Chan_2021_ICCV} & \cmark & 85.47 & 15.00 & 49.21 & 39.51 & 28.72 & 85.07 & 0.75  & 47.87 & 62.64 & 48.51 & 77.90 & 9.70  & 45.90 & 63.06 & 49.92\\
Mask2Anomaly~\cite{Rai_2023_ICCV}      & \cmark & 88.72 & 14.63 & 55.28 & 51.68 & 47.16 & 93.22 & 0.20  & 55.72 & 75.42 & 68.15 & --    & --    & --    & --    & --   \\
EAM~\cite{Grcic_2023_CVPR}               & \cmark & 93.75 & 4.09  & 67.09 & 53.77 & 60.86 & 92.87 & 0.52  & 65.86 & 76.50 & 75.58 & --    & --    & --    & --    & --   \\
RbA~\cite{nayal2023ICCV}               & \cmark & 94.46 & 4.60  & 64.93 & 47.51 & 51.87 & 95.12 & 0.08  & 54.34 & 59.08 & 57.44 & --    & --    & --    & --    & --   \\

\cmidrule(lr){1-17}

ODIN~\cite{liang2018enhancing}              & \xmark & 33.06 & 71.68 & 19.53 & 17.88 & 5.15  & 22.12 & 15.28 & 21.62 & 18.50 & 9.37  & 52.93 & 30.04 & 39.79 & 49.33 & 34.53\\
JSRNet~\cite{Vojir_2021_ICCV}            & \xmark & 33.64 & 43.85 & 20.20 & 29.27 & 13.66 & 28.09 & 28.86 & 18.55 & 24.46 & 11.02 & 74.17 & 6.59  & 34.28 & 45.89 & 35.97\\
PEBAL~\cite{Tian2021PixelwiseEA}             & \xmark & 49.14 & 40.82 & 38.88 & 27.20 & 14.48 & 4.98  & 12.68 & 29.91 & 7.55  & 5.54  & --    & --    & --    & --    & --   \\
Image Resynthesis~\cite{Lis_2019_ICCV} & \xmark & 52.28 & 25.93 & 39.68 & 10.95 & 12.51 & 37.71 & 4.70  & 16.61 & 20.48 & 8.38  & 57.08 & 8.82  & 27.16 & 30.69 & 19.17\\
Road Inpainting~\cite{lis2023detecting}   & \xmark & --    & --    & --    & --    & --    & 54.14 & 47.12 & 57.64 & 39.50 & 36.01 & 82.93 & 35.75 & 49.21 & 60.67 & 52.25\\
NFlowJS~\cite{grcic2023dense}           & \xmark & 56.92 & 34.71 & 36.94 & 18.01 & 14.89 & 85.55 & 0.41  & 45.53 & 49.53 & 50.36 & 89.28 & 0.65  & 54.63 & 59.74 & 61.75\\
ObsNet~\cite{besnier2021trigger}            & \xmark & 75.44 & 26.69 & 44.22 & 52.56 & 45.08 & --    & --    & --    & --    & --    & --    & --    & --    & --    & --   \\
DaCUP~\cite{Vojir_2023_WACV}             & \xmark & --    & --    & --    & --    & --    & 81.50 & 1.13  & 37.68 & 60.13 & 46.01 & 81.37 & 7.36  & 38.34 & 67.29 & 51.14\\
RbA~\cite{nayal2023ICCV}               & \xmark & 86.13 & 15.94 & 56.26 & 41.35 & 42.04 & 87.85 & 3.33  & 47.44 & 56.16 & 50.42 & --    & --    & --    & --    & --   \\
CSL~\cite{Zhang_Li_Qi_Yang_Ahuja_2024}  & \xmark & 80.08 &7.16 &46.46 & 50.02 &50.39 & 87.10 &0.67 &44.70&53.13&51.02& --    & --    & --    & --    & --   \\
cDNP~\cite{Galesso_2023_ICCV}              & \xmark & 88.90 & 11.42 & 50.44 & 29.04 & 28.12 & --    & --    & --    & --    & --    & --    & --    & --    & --    & --   \\
\pixood ({\bf Ours}) & \xmark & 68.88 & 54.33 & 44.15 & 24.32 & 19.82 & 88.90 & 0.30  & 42.68 & 57.49 & 50.82 & 85.07 & 4.46 & 30.18 & 78.47 & 44.41\\

\bottomrule
\end{tabular}
}
\label{tab:smiyc_results_all}
\end{table*}


\end{document}